\documentclass{article}

\usepackage{arxiv}

\usepackage[utf8]{inputenc} 
\usepackage[T1]{fontenc}    
\usepackage{hyperref}       
\usepackage{url}            
\usepackage{booktabs}       
\usepackage{amsfonts}       
\usepackage{nicefrac}       
\usepackage{microtype}      
\usepackage{lipsum}		
\usepackage{graphicx}
\usepackage{natbib}
\usepackage{doi}

\usepackage{xcolor}         
\usepackage{utfsym}
\usepackage{multicol}
\usepackage{multirow}
\usepackage{makecell}
\usepackage{colortbl}

\usepackage{xcolor}

\usepackage{amsmath}
\usepackage{amsthm}
\newtheorem{definition}{\textbf {Definition}}

\usepackage{listings}
\usepackage{xcolor}

\usepackage[most]{tcolorbox}

\lstdefinestyle{prompt}{
  basicstyle=\ttfamily\small,
  columns=fullflexible,
  breaklines=true,
  breakatwhitespace=true,
  keepspaces=true,
  showstringspaces=false,
  frame=none,
  aboveskip=0pt,
  belowskip=0pt,
  language={},
}

\title{Text-ADBench: Text Anomaly Detection Benchmark Based on LLM Embeddings}

\author{Feng Xiao \\
	The Chinese University of Hong Kong, Shenzhen\\
	Shenzhen, China\\
	\texttt{fengxiao1@link.cuhk.edu.cn} \\
	\And
	Jicong Fan\thanks{Corresponding Author} \\
	The Chinese University of Hong Kong, Shenzhen\\
	Shenzhen, China\\
	\texttt{fanjicong@cuhk.edu.cn} \\
}

\renewcommand{\shorttitle}{\textit{arXiv} Template}

\hypersetup{
pdftitle={A template for the arxiv style},
pdfsubject={q-bio.NC, q-bio.QM},
pdfauthor={David S.~Hippocampus, Elias D.~Striatum},
pdfkeywords={First keyword, Second keyword, More},
}

\begin{document}
\maketitle

\begin{abstract}
Text anomaly detection is a critical task in natural language processing (NLP), with applications spanning fraud detection, misinformation identification, spam detection and content moderation, etc. Despite significant advances in large language models (LLMs) and anomaly detection algorithms, the absence of standardized and comprehensive benchmarks for evaluating the existing anomaly detection methods on text data limits rigorous comparison and development of innovative approaches. This work performs a comprehensive empirical study and introduces a benchmark for text anomaly detection, leveraging embeddings from diverse pre-trained language models across a wide array of text datasets. Our work systematically evaluates the effectiveness of embedding-based text anomaly detection by incorporating (1) early language models (GloVe, BERT); (2) multiple LLMs (LLaMA-2, LLaMA-3, Mistral, OpenAI embedding models (small, ada, large)); (3) multi-domain text datasets (news, social media, scientific publications); (4) comprehensive evaluation metrics (AUROC, AUPRC). 
Our experiments reveal a critical empirical insight: embedding quality significantly governs anomaly detection efficacy, and deep learning-based approaches demonstrate no performance advantage over conventional shallow algorithms (e.g., KNN, OCSVM) when leveraging LLM-derived embeddings.
In addition, we observe strongly low-rank characteristics in cross-model performance matrices, which enables an efficient strategy for rapid model evaluation (or embedding evaluation) and selection in practical applications. Furthermore, by open-sourcing our benchmark toolkit that includes all embeddings from different models and code, this work provides a foundation for future research in robust and scalable text anomaly detection systems. Our code repository is at~\url{https://github.com/jicongfan/Text-Anomaly-Detection-Benchmark}.
\end{abstract}


\section{Introduction}

Anomaly detection (AD) \cite{10.1145/1541880.1541882,aggarwal2017introduction,ruff2021unifying}, broadly defined as the task of identifying patterns that deviate significantly from the majority, is a cornerstone of data analysis and decision-making in different fields. In the past few decades, researchers have achieved significant progress in medical diagnosis \cite{kononenko2001machine,richens2020improving}, intrusion detection~\cite{mukherjee2002network,khraisat2019survey} in cybersecurity, fraud detection~\cite{bolton2002statistical,abdallah2016fraud}
in finance, and fault detection~\cite{isermann2005model,fan2017autoencoder,fan2021kernel} in industry. While these AD methods commonly excel in structured data type, such as time-series, tabular data, images and videos, text-based anomaly detection (see the formal definition given by Definition \ref{def_text_ad} and the corresponding examples) presents challenges due to the unstructured, high-dimensional nature of language, complex and diverse anomalies, where anomalies may manifest as subtle semantic inconsistencies, syntactic irregularities, or contextual deviations. 

Due to the inherent characteristics of human language, text representation techniques have become a fundamental component of natural language processing (NLP). Early text representation techniques ~\cite{harris1954distributional,sparck1972statistical,aizawa2003information} relied heavily on handcrafted features, such as term frequency (TF) and inverse document frequency (IDF), which aimed to quantify the importance of words within documents. These methods, while simple and computationally efficient, treated text as unordered collections of words (i.e., bag-of-words models), failing to capture syntactic structure or semantic relationships. Subsequent developments introduced vector space models, such as Latent Semantic Analysis (LSA)~\cite{blei2003latent} and Latent Dirichlet Allocation (LDA)~\cite{dumais2004latent}, incorporating co-occurrence patterns and topic distributions to produce more meaningful document embeddings. However, these techniques still struggled to represent contextual meaning and polysemy, limiting their expressiveness in downstream tasks. A major breakthrough occurred with the rise of neural word embeddings, particularly Word2Vec~\cite{mikolov2013efficient}, GloVe~\cite{pennington-etal-2014-glove}, and FastText~\cite{joulin2016bag}, which mapped words into dense, low-dimensional vectors based on their distributional properties in large corpora. These models significantly improved the semantic fidelity of word representations, but they remained static, assigning each word a single representation regardless of its context. To address this limitation, researchers introduced contextualized word embeddings, such as ELMo~\cite{peters2018deepcontextualizedwordrepresentations}, BERT~\cite{devlin2019bert} and GPT~\cite{radford2018improving}, which generated dynamic representations conditioned on surrounding text. These models marked a paradigm shift in NLP by allowing representations to reflect both syntax and semantics within a specific context. With the advent of large language models (LLMs) such as GPT-series, LLaMA-series and etc., leveraging transformer architectures and pretraining on large-scale text data, text representation based on LLMs exhibits remarkable performance in a wide range of language understanding and generation tasks~\cite{naveed2023comprehensive, patil2024review}.

However, the integration of LLM-based representation learning into text anomaly detection remains insufficiently explored, and there is a notable lack of comprehensive analysis and systematic comparison across conventional shallow AD algorithms~\cite{scholkopf1999support,breunig2000lof,liu2008isolation}, deep learning-based AD methods~\cite{ruff2018deep, 8836638, fu2024dense}, and tailored text AD approaches~\cite{ruff2019self,manolache2021date}. This gap has hindered the progress and development of text anomaly detection techniques. We note that two recent studies~\cite{li2024nlp,cao2025tad} provided preliminary comparisons and valuable insights. However, these works exhibit several limitations: (1) limited comparative scope, primarily focusing on shallow anomaly detection algorithms; (2) lack of analysis on different pooling strategies when leveraging open-sourced LLM-based text embeddings, despite their significant impact on representation quality; (3) a narrow evaluation framework relying on a single performance metric; and (4) lack of open-source embeddings and code, limiting reproducibility and further research.

\begin{figure}
    \centering
    \includegraphics[width=0.90\linewidth]{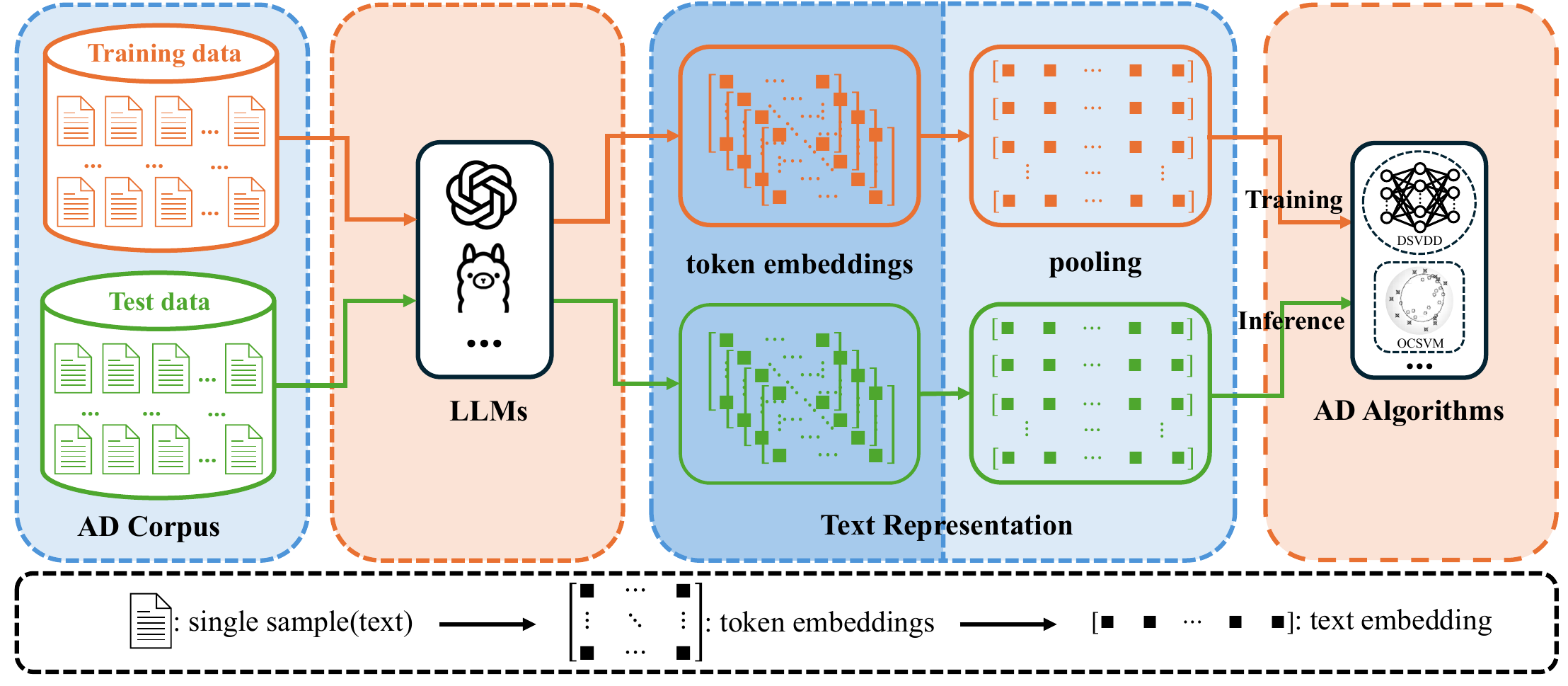}
    \caption{Flowchart of Text-ADBench.}
    \label{fig-flow_chart}
\end{figure}

To address these gaps, this work introduces Text-ADBench, a comprehensive benchmark for text anomaly detection based on embeddings derived from LLMs. Specifically, we construct a large number of two-stage text AD pipelines. First, we generate the text embedding using a diverse suite of language models, including early language models (GloVe-6B, BERT), open-source LLMs (LLaMA2-7B, LLaMA3-8B, Mistral-7B) and OpenAI’s text-embedding models (small, ada, large). To aggregate sequential token embeddings into a single vector representation, we utilize three pooling strategies, including ``mean'', ``end-of-sequence (EOS) token'', and ``weighted mean''. Second, we develop anomaly detection tasks by applying these embeddings to a range of AD methods, encompassing both shallow and deep learning-based techniques. The workflow of Text-ADBench is illustrated in Figure~\ref{fig-flow_chart}. First, the text data are fed into pretrained LLMs tailored for text embedding, from which token-level representations are obtained. Subsequently, different text embeddings are derived using various pooling strategies. Finally, these text embeddings are utilized for anomaly detection using a range of AD algorithms. Additionally, we incorporate two specialized text AD methods (CVDD~\cite{ruff2018deep} and DATE~\cite{manolache2021date}) into our comparative analysis. Our experiments are conducted on eight real-world text datasets. The main contributions of this work are summarized as follows.

\begin{itemize}
    \item We present a comprehensive benchmark for text anomaly detection, addressing the existing gap in leveraging large language models (LLMs) embeddings for this task. Our framework incorporates diverse language models, multiple pooling strategies, a wide range of anomaly detection methods, and comprehensive evaluation metrics
    \item We conduct further analysis of the results and identify a low-rank property in performance. This finding has two important implications: (1) the detection performance of novel text datasets or AD methods can be reliably predicted using only a subset of performance measurements, and (2) this property enables an efficient strategy for rapid model evaluation (embedding evaluation) and selection in practical applications.
    \item We release Text-ADBench at \url{https://github.com/jicongfan/Text-Anomaly-Detection-Benchmark} that includes all data, embeddings, and code used in our experiments. We provide an integrated and easy-to-use framework that allows researchers to reproduce results or to evaluate novel embeddings and methods, which would foster the development of these fields.  
\end{itemize}

This work serves as a foundational resource for both researchers and practitioners, advancing the text AD field through its methodological rigor and holistic evaluation. By open-sourcing our benchmark framework, precomputed embeddings, we aim to catalyze further research into hybrid pooling strategies, domain adaptation, and efficient LLM utilization for text anomaly detection.

The remainder of this paper is organized as follows: Section 2 reviews related work; Section 3 formulates the problem of text anomaly detection; Section 4 details the specific settings including datasets, LLMs, embedding, AD methods. Section 5 presents experimental results and basic analysis; Section 6 explores the low-rank property of performance; and Section 7 concludes this work and provide future directions.

\section{Related Work}

\subsection{Unsupervised Anomaly Detection}

Unsupervised Anomaly Detection (UAD) is a fundamental task in data analysis and decision-making, with broad applications across various domains, such as fault detection in industry, fraud detection in finance, medical diagnosis, and intrusion detection in cybersecurity. Consequently, a flood of unsupervised anomaly detection methods have been proposed in recent decades~\cite{scholkopf1999support, breunig2000lof, liu2008isolation, ruff2018deep, deecke2019image,qiu2021neural, fu2024dense, xiao2025unsupervised, dai2025autouad}.
Generally, these UAD methods are grounded in distinct assumptions about data distribution~\cite{aggarwal2017introduction, pang2021deep, ruff2021unifying}. For instance, anomalies may reside in low-density regions or exhibit deviations in feature space compared to normal instances. The performance of these methods is often contingent upon the degree of alignment between the input data and the underlying assumptions. Roughly, these methods can be organized into two categories, including conventional machine learning-based shallow algorithms and deep learning-based methods \cite{pang2021deep}. Shallow anomaly detection algorithms~\cite{scholkopf1999support, breunig2000lof, liu2008isolation, li2022ecod} typically have a straightforward learning process and exhibit superior interpretability. For instance, OCSVM~\cite{scholkopf1999support} learns a maximum-margin hyperplane in a kernel space to separate normal data from the origin. Isolation Forest~\cite{liu2008isolation} builds an ensemble of trees to isolate data points, where anomalies are identified by their shorter average path lengths on these trees. KNN~\cite{ramaswamy2000efficient} defines the anomaly score of a sample as its distance to the k-th nearest neighbor, based on the assumption that anomalies reside in sparser regions of the feature space compared to normal data. Deep learning-based anomaly detection methods~\cite{schlegl2017unsupervised, dagmm, ruff2018deep, pang2019deep, perera2019ocgan, goyal2020drocc, fan2021kernel, qiu2021neural, cai2022perturbation, xu2023deep, fu2024dense, ye2025drl} often achieve superior detection performance, particularly in high-dimensional data settings. For instance, AutoEncoder (AE)~\cite{hinton2006reducing} minimizes the reconstruction error on normal data. It assumes that abnormal data cannot be reconstructed well when an autoencoder is trained only on normal data. Naturally, reconstruction error is used as an anomaly score. Deep SVDD~\cite{ruff2018deep} learns a minimum-radius hypersphere to encompass all the normal data while the unseen abnormal data are expected to fall outside, where its anomaly score is defined by the distance to the hypersphere's center. The PLAD proposed by \cite{cai2022perturbation} trains a perturbator to perturb normal data and a classifier to distinguish normal data and perturbed data, leading to an effective decision boundary for anomaly identification. 

Note that in addition to UAD methods, there are also a few semi-supervised AD methods~\cite{hendrycks2018deep, pang2018learning, akcay2019ganomaly, ruff2019deep, zhou2021feature, pang2023deep, xiao2025semi} that can utilize labeled anomalies to improve the detection performance, since in some scenarios, a few known anomalous samples, though scarce, are available during the training stage.

\subsection{Text Representation}

Text representation is a fundamental technique of natural language processing (NLP) that transforms unstructured textual data into structured numerical formats, thereby enabling machine learning models to process and analyze linguistic information effectively. The evolution of text representation techniques has undergone several significant stages.


Early text representation techniques primarily relied on statistical and frequency-based approaches. The Bag-of-Words (BOW) model~\cite{harris1954distributional} and Term Frequency-Inverse Document Frequency (TF-IDF)~\cite{sparck1972statistical} represented text as vectors of word frequencies, thereby only preserving the statistical lexical information but neglecting word order. As a result, the BOW model fails to capture semantic relationships between words and lacks contextual understanding. To address the limitations of the BOW model, distributed representations~\cite{mikolov2013efficient, pennington2014glove} were developed, which encode words as dense vectors in a continuous space. For instance, Word2Vec~\cite{mikolov2013efficient} introduced two architectures, Skip-gram and Continuous Bag-of-Words (CBOW), that capture semantic and syntactic relationships by predicting word contexts. Similarly, GloVe (Global Vectors for Word Representation)~\cite{pennington2014glove} integrated global co-occurrence statistics with local context window-based learning. However, despite their advancements, Word2Vec and GloVe produce static and context-free word embeddings, which are unable to effectively handle polysemy and out-of-vocabulary (OOV) words. In 2017, the introduction of Transformer architecture~\cite{vaswani2017attention} revolutionized text representation by generating context-dependent embeddings. Subsequent models, such as GPT (Generative Pretrained Transformer)~\cite{radford2018improving} and BERT (Bidirectional Encoder Representations from Transformers)~\cite{devlin2019bert}, leveraged the self-attention mechanism to produce dynamic word representations conditioned on surrounding text, markedly improving performance on downstream NLP tasks. With the advent of LLMs (Large Language Models), recently, the NLP community has begun adopting decoder-only LLMs for text embedding~\cite{wang2023improving, behnamghader2024llm2vec}.

\subsection{Text Anomaly Detection}

In the field of natural language processing, anomaly detection tasks are employed to identify harmful content, spam reviews, and abusive language. A straightforward strategy is to construct pipeline (two-stage) methods by integrating text representation techniques with existing anomaly detection methods. Xu et al.~\cite{xu2023comparative} systematically estimated such combinations based on SBERT~\cite{reimers2019sentence}. Additionally, several specialized text anomaly detection algorithms~\cite{ruff2019self, manolache2021date, das2023few} were proposed in recent years. For instance, CVDD~\cite{ruff2018deep}, a self-attention-based model for unsupervised text anomaly detection method, minimizes the weighted cosine distance between the attention-weighted text embeddings and ``context vectors'', where it employs self-attention heads to generate ``context vectors'' that capture diverse semantic themes in normal data. Each head produces an orthogonal attention matrix, ensuring independent feature extraction. During the inference stage, anomalies are identified by averaging the cosine distances between the new sample’s embeddings and all ``context vectors''. DATE~\cite{manolache2021date} trains Transformers using masked language modeling (MLM) and two tailored pretext tasks to capture contextual anomalies. FATE~\cite{das2023few}, a semi-supervised text anomaly detection method, employs a contrastive objective to adjust anomaly scores. This approach ensures that normal samples cluster around a reference score while anomalies deviate significantly. Additionally, FATE leverages attention mechanisms and multi-instance learning to capture distinctive anomaly behaviors. More recently, Pantin and Marsala~\cite{pantin2024robust} introduced ERLA, a robust ensemble-based anomaly detection method utilizing autoencoders, where each autoencoder incorporates a local robust subspace recovery projection of the original data in its encoding embedding, thereby leveraging the geometric properties of the k-nearest neighbors to optimize subspace recovery and identify anomalous patterns in textual data.

\subsection{Anomaly Detection based on LLMs}


The remarkable success of large language models (LLMs) in natural language processing has spurred their adaptation to anomaly detection (AD) tasks~\cite{su2024large} for further improvement of detection accuracy via their advanced semantic understanding. Current LLM-based AD approaches can be categorized into three paradigms: \textbf{\textit{(i) prompt-based methods}}~\cite{yang2024ad,dong2024can} directly employ pretrained LLMs to perform few-shot or zero-shot learning by designing specialized prompt functions, and \textbf{\textit{(ii) fine-tuning methods}}~\cite{gu2024anomalygpt,li2024anomaly}  adapt general-purpose LLMs into domain-specific anomaly detectors by fine-tuning them on task-relevant datasets, thereby tailoring their capabilities to the nuances of AD, and \textbf{\textit{(iii) embedding-based methods}}~\cite{li2024nlp,cao2025tad} regard LLMs as feature extractors, generating high-dimensional embeddings that capture high-quality semantic and contextual patterns and then leveraging existing AD algorithms (e.g., OCSVM, Isolation Forest) to detect deviations from normal pattern based on these embeddings.

\subsection{Existing Benchmarks for Text Anomaly Detection}

There are some notable works~\cite{xu2023comparative,bejan2023ad,li2024nlp,yang2024ad,cao2025tad} that take effort to text anomaly detection. For instance, Xu et al.~\cite{xu2023comparative} evaluated 22 AD algorithms on 17 text corpora and mainly analyzed the performance effects of weak supervised signals. To the best of our knowledge, the first text anomaly detection benchmark based on LLMs is NLP-ADBench~\cite{li2024nlp}, where Li et al. evaluated 11 AD algorithms based on embeddings of BERT and OpenAI (text-embedding-3-large) across 8 text corpora. To clearly delineate the different emphases and advantages of existing benchmarks and to further compare them with Text-ADBench, we provide a statistical comparison in Table~\ref{tab-benchmarks-com}. In summary, this benchmark emphasizes the following three aspects. 
\begin{itemize}
    \item 
First,  while previous works~\cite{xu2023comparative, bejan2023ad} mainly focus on the embeddings derived from early language models, recent works~\cite{li2024nlp, cao2025tad} have begun to incorporate embeddings from LLMs, albeit with a limited scope. Considering the rapid advancement of the representation capabilities of LLMs, we compare 33 distinct embeddings for each dataset by integrating embedding models with pooling strategies. Notably, previous works do not compare the impact of different pooling strategies. 
\item Second, this benchmark identifies a low-rank property in performance matrices and demonstrates that the detection performance of novel text datasets or AD methods can be effectively predicted based on historical performance measurements. This finding enables an efficient strategy for rapid model evaluation (or embedding evaluation) or selection based on the results of this benchmark. 
\item More importantly, we release all resources used in this benchmark including all original text datasets, embeddings and code. These open-access resources ensure researchers and practitioners can easily and efficiently reproduce our results, estimate novel datasets or AD algorithms, thereby fostering progress in the field. Additionally, the released embeddings can be leveraged for other downstream tasks.
\end{itemize}
\begin{table}[h!]
    \centering
    \caption{Comparison among Text-ADBench and existing benchmarks, where ``DL'' and``TFT'' refers to ``Deep Learning'' and ``Tailored for Text'', respectively. Note that the term ``\# Emb.'' denotes the number of distinct embeddings derived from each text dataset.}
    \label{tab-benchmarks-com}
    \resizebox{\linewidth}{!}{
    \begin{tabular}{l|ccccc|ccc|cc|cc|cc}
    \toprule
    \multirow{2}{*}{\textbf{Benchmark}} & \multicolumn{5}{c|}{\textbf{Coverage} (numbers)} & \multicolumn{3}{c|}{\textbf{AD Algorithm Type}} & \multicolumn{2}{c|}{\textbf{Representation}} & \multicolumn{2}{c|}{\textbf{Open Source}} & \multicolumn{2}{c}{\textbf{Performance Matrices}} \\
    & Datasets & Algo. & LLMs & Emb. & Metrics & Shallow & DL & TFT & Early LM & Pooling & Code & Embedding & Analysis & Prediction \\
    \midrule
    \text{Xu et al.}~\cite{xu2023comparative} (2023) & 22 & 17 & 0 & 2 &  1 &  \usym{1F5F8} & \usym{1F5F8} & \usym{2717} & \usym{1F5F8} & \usym{2717} & \usym{2717} & \usym{2717} & \usym{2717} & \usym{2717}\\
    \text{Bejan et al.}~\cite{bejan2023ad} (2023) & 7 & 4 & 0 & 2 & 2 & \usym{1F5F8} & \usym{2717} & \usym{1F5F8} & \usym{1F5F8} & \usym{2717} & \usym{1F5F8} & \usym{2717} & \usym{2717} & \usym{2717}\\
    \text{Li et al.}~\cite{li2024nlp} (2024) & 8 & 11 & 1 & 2 & 1 & \usym{1F5F8} & \usym{1F5F8} & \usym{1F5F8} & \usym{1F5F8} & \usym{2717} & \usym{1F5F8} & \usym{2717} & \usym{2717} & \usym{2717}\\
    \text{Yang et al.}~\cite{yang2024ad} (2024) & 5 & 10 & 2 & - & 2 & \usym{1F5F8} & \usym{1F5F8} & \usym{1F5F8} & \usym{2717} & \usym{2717} & \usym{1F5F8} & \usym{2717} & \usym{2717} & \usym{2717} \\
    \text{Cao et al.}~\cite{cao2025tad} (2025)& 6 & 8 & 5 & 8 & 1 & \usym{1F5F8} & \usym{2717} & \usym{2717} & \usym{1F5F8} & \usym{2717} & \usym{2717} & \usym{2717} & \usym{2717} & \usym{2717}\\
    \text{Maia et al.}~\cite{maia2025learning} (2025) & 6 & 12 & 0 & 6 & 1 & \usym{1F5F8} & \usym{1F5F8} & \usym{2717} & \usym{1F5F8} &  \usym{2717} & \usym{1F5F8} &  \usym{2717} & \usym{2717} &  \usym{2717}\\
    \midrule
    \text{Text-ADBench (ours)} & 12 & 12 & \textcolor{red}{12} & \textcolor{red}{33} & 2 & \usym{1F5F8} & \usym{1F5F8} & \usym{1F5F8} & \usym{1F5F8} & \textcolor{red}{\usym{1F5F8}} & \usym{1F5F8} & \textcolor{red}{\usym{1F5F8}} & \textcolor{red}{\usym{1F5F8}} & \textcolor{red}{\usym{1F5F8}}\\
    \bottomrule
    \end{tabular}}
\end{table}

\section{Problem Formulation}
Text anomaly detection, considered in this work, aims to identify the text instances that deviate significantly from the majority. The following presents a formal definition.
\begin{definition}[Text anomaly detection]\label{def_text_ad}
    Let $\mathcal{C} = \{s_1, s_2, \cdots, s_n\}$ be a corpus of $n$ textual sequences, where all or most sequences are in some unknown normal condition or pattern $P$. Text anomaly detection aims to learn a detector $f$ from $\mathcal{C}$ that can determine whether a new textual sequence $s_{\text{new}}$ is in $P$ or not.
\end{definition}
For instance, if $\mathcal{C}$ is composed of emails, a textual sequence containing spam information should be detected as abnormal. Another example, if $\mathcal{C}$ consists of movie reviews, a review with abusive content would also be detected as abnormal. 

In this work, we split the text anomaly detection into two stages. In the first stage, we extract text embeddings based on early language models (GloVe and BERT) and LLMs. Subsequently, in the second stage, we design a general unsupervised anomaly detection task on the text embeddings. Before delving into the details, we first introduce the notation convention as follows.

\begin{enumerate}
    \item \textbf{Text Representation based on Embedding Models}
    \quad Let $\mathbf{M}_\text{emb}$ be a language model (e.g., BERT and LLaMA3). We obtain the embeddings $ \mathcal{X}=\{\mathbf{x}_1, 
\mathbf{x}_2, \cdots, \mathbf{x}_n\}$ of $\mathcal{C}$ as
\begin{equation}
    \mathbf{x}_i = Pooling\big(\mathbf{M}_\text{emb}(s_i)\big), ~~i=1,2,\cdots, n,
\end{equation}
where $Pooling(\cdot)$ aims to aggregate token-level embeddings of the sequence $s_i$ into a single vector $\mathbf{x}_i \in \mathbb{R}^d$, where $d$ represents the embedding dimension of the language model.

    \item \textbf{Unsupervised Anomaly Detection}
\quad We assume that $\mathcal{X}$ is drawn from an unknown distribution $\mathcal{D}_{\mathbf{x}} \subseteq \mathbb{R}^d$. A point $\mathbf{x} \in \mathbb{R}^d$ is deemed to be anomalous if $\mathbf{x} \notin \mathcal{D}_{\mathbf{x}}$.  Then, the goal of the unsupervised AD is to obtain a decision function $h_{\text{UAD}}: \mathbb{R}^d \rightarrow \{0, 1\}$ by utilizing only $\mathcal{X}$, such that $h_{\text{UAD}}(\mathbf{x}) = 0$ if $\mathbf{x} \in \mathcal{D}_{\mathbf{x}}$ and $h_{\text{UAD}}(\mathbf{x}) = 1$ if $\mathbf{x} \notin \mathcal{D}_{\mathbf{x}}$. The primary difference among AD methods lies in the design of the decision function $f(\cdot)$. 
\end{enumerate}
Based on the two stages above, the detector $f$ can be formulated as
\begin{equation}
    f(s):=h_{\text{UAD}}\left(Pooling\big(\mathbf{M}_\text{emb}(s)\big)\right)
\end{equation}
The value of $f(s_{\text{new}})$ can determine whether the textual sequence $s_{\text{new}}$ is normal or anomalous.

In this work, we consider the combinations of different embedding models, different pooling operations, and different UAD algorithms, leading to a comprehensive evaluation of key techniques for text anomaly detection.





\section{Benchmark Settings}
In this section, we will introduce the detailed experimental preparations, including dataset splitting, embedding models, and the anomaly detection methods used in this benchmark.

\subsection{Datasets}

Different from anomaly detection tasks on images, tabular data, and time series, where there are standard datasets with normal and abnormal samples defined clearly, text anomaly detection presents unique challenges. The richness of natural language allows anomalies to manifest in diverse forms, such as rare topics, unusual language styles, irregular syntax or grammar, and harmful or abusive content. 
We observe that existing textual AD benchmarks~\cite{xu2023comparative,bejan2023ad,li2024nlp,yang2024ad,cao2025tad} consistently employ the NLP classification datasets. Aligning with this established methodology, our benchmark utilizes 8 text classification datasets from various NLP domains to construct 14 specialized Text-AD datasets. The statistical information of the Text-AD datasets are shown in Table~\ref{tab-datasets}.

\begin{table}[h!]
    \centering
    \caption{Statistical information of the Text-AD datasets.}
    \label{tab-datasets}
    \begin{tabular}{c|cccc}
    \toprule
    \textbf{Text-AD Dataset} & \textbf{Domain} & \# \textbf{Training Samples} & \# \textbf{Test Samples} & \textbf{Anomaly Ratio} \\
    \midrule
    20Newsgroups & News & 10,419 & 7,876 & 0.12 \\
    Reuters21578 & News & 4,435 & 4,723 & 0.45 \\
    IMDB & Movie Review & 10,000 & 40,000 & 0.625 \\
    SST2 & Movie Review & 10,000 & 58,078 & 0.51 \\
    SMS-Spam & Phone Message & 3,000 & 2,534 & 0.29 \\
    Enron & Email & 10,000 & 21,924 & 0.31\\
    WOS & Paper Abstract & 28,918 & 18,065 & 0.31 \\
    DBpedia14-0 & Wikipedia Term & 20,000 & 44,999 & 0.44\\
    DBpedia14-1 & Wikipedia Term & 20,000 & 44,999 & 0.44\\
    DBpedia14-2 & Wikipedia Term & 20,000 & 44,999 & 0.44\\
    DBpedia14-3 & Wikipedia Term & 20,000 & 45,000 & 0.44\\
    DBpedia14-4 & Wikipedia Term & 20,000 & 44,997 & 0.44\\
    \bottomrule
    \end{tabular}
\end{table}

The specific splitting and descriptions of each dataset are summarized below.

\begin{itemize}

\item \textbf{20Newsgroups}\footnote{http://qwone.com/~jason/20Newsgroups/}~\cite{Lang95} is a collection of newsgroup documents. It is organized into 20 different newsgroups, each corresponding to a different topic. Some of the newsgroups are very closely related to each other (e.g., ``comp.sys.ibm.pc.hardware / comp.sys.mac.hardware''), while others are highly unrelated (e.g ``misc.forsale / soc.religion.christian''). For constructing the text-AD dataset, class ``misc.forsale'' is regarded as an abnormal class, and the remaining classes are regarded as normal. 

\item \textbf{Reuters21578}\footnote{https://raw.githubusercontent.com/nltk/nltk\_data/gh-pages/packages/corpora/reuters.zip}~\cite{Reuters-21578} is a collection of documents that appeared on Reuters newswire in 1987. It is organized into 59 different groups with 21578 samples in the original version. In this work, we use the ApteMod version provided in NLTK Corpora\footnote{https://www.nltk.org/nltk\_data/} and classes ``earn'' and ``acq'' are regarded as normal, and the remaining are regarded as abnormal.

\item \textbf{IMDB}\footnote{http://ai.stanford.edu/~amaas/data/sentiment/}~\cite{maas-EtAl:2011:ACL-HLT2011} is a dataset for binary sentiment (``pos'' and ``neg'') classification containing substantially more data than previous benchmark datasets. For constructing the text-AD dataset, class ``pos'' is regarded as normal, and ``neg'' is regarded as the abnormal class.

\item \textbf{SST2}\footnote{https://huggingface.co/datasets/stanfordnlp/sst2}~\cite{socher-etal-2013-recursive} is a corpus with fully labeled parse trees that allows for a complete analysis of the compositional effects of sentiment in language. The corpus is based on the dataset introduced by Pang and Lee (2005) and consists of 11,855 single sentences extracted from movie reviews. It is a binary sentiment classification dataset. In this benchmark, class ``1-(positive)'' is regarded as the normal class and class ``0-(negative)'' is regarded as the abnormal class.

\item \textbf{SMS-Spam}\footnote{https://huggingface.co/datasets/ucirvine/sms\_spam}~\cite{Almeida2011SpamFiltering} is a public set of SMS labeled messages that have been collected for mobile phone spam research. It has one collection composed by 5,574 English, real and non-encoded messages, tagged according being legitimate or spam. Naturally, legitimate messages are regarded the normal samples, and spam messages are regarded as abnormal samples.

\item \textbf{Enron}\footnote{https://huggingface.co/datasets/Hellisotherpeople/enron\_emails\_parsed} contains a total of about 0.5M messages from about 150 users, mostly senior management of Enron. We counted the entire dataset and found the two email accounts (``kay.mann@enron.com'', ``vince.kaminski@enron.com'') with the most emails under them, both greater than 10,000, as well as many accounts with just 1 email under them. Based on this observation, we use the emails from accounts (``kay.mann@enron.com'', ``vince.kaminski@enron.com'') as normal samples and the emails from the accounts with only 1 email as abnormal samples.

\item \textbf{WOS}\footnote{https://huggingface.co/datasets/river-martin/web-of-science-with-label-texts}~\cite{kowsari2017HDLTex} is Web of Science Dataset (WOS-46985). This dataset contains the abstracts of 46,985 published papers that have 7 parent categories, including ``Computer Science (CS), Electrical Engineering (ECE), Psychology, Mechanical Engineering (MAE), Civil Engineering (Civil), Medical Science (Medical), and Biochemistry''.
For constructing the text-AD dataset, class ``Psychology'' is regarded as the normal class, and the remaining classes are regarded as abnormal.
\item \textbf{DBpedia14}\footnote{https://huggingface.co/datasets/fancyzhx/dbpedia\_14}~\cite{NIPS2015_250cf8b5} is constructed by picking 14 non-overlapping classes from DBpedia 2014. Each class contains 40,000 training samples and 5,000 testing samples. We construct five text-AD datasets (DBpedia14-0, DBpedia14-1, DBpedia14-2, DBpedia14-3, DBpedia14-4) based on the original dataset. Specifically, for DBpedia14-0, we use class 0 as the normal class, and the samples from the test set of classes [1, 2, 3, 4] are regarded as abnormal samples. For DBpedia14-1, we use class 1 as the normal class, and the samples from the test set of classes [0, 2, 3, 4] are regarded as abnormal samples. The text-AD datasets (DBpedia14-2, DBpedia14-3, DBpedia14-4) have the same construction process. 

\end{itemize}

\subsection{Embedding Models}

Text embedding is widely recognized as the foundational and essential step in NLP tasks, as it transforms the semantic content of natural language into a structured vector representation. Over the past several years, the emergence of innovative embedding techniques has consistently propelled the field of NLP forward, significantly advancing the development and applications of NLP techniques. 

For instance, researchers at Stanford University advanced the field of word embeddings with the introduction of GloVe (Global Vectors for Word Representation)~\cite{pennington-etal-2014-glove}. GloVe enhanced the capabilities of Word2Vec by incorporating global statistical information across the entire corpus to generate word vectors, which enabled a more nuanced semantic understanding by integrating both local context windows and global corpus statistics. The landscape of NLP was further revolutionized in 2017 with the introduction of the transformer architecture~\cite{vaswani2017attention}, which introduced the self-attention mechanism. Building on this breakthrough, BERT(Bidirectional Encoder Representation from Transformers)~\cite{devlin2019bert}, released in 2018, pioneered context-dependent word embeddings. Unlike earlier models such as Word2Vec and GloVe, which produced static, context-free embeddings, BERT leveraged a transformer-based architecture to create dynamic representations that capture the meaning of words based on their surrounding context, considering both preceding and succeeding words within a sentence. With the advent of LLMs (Large Language models), only recently, the NLP community has started to adopt decoder-only LLMs for text embedding~\cite{behnamghader2024llm2vec}.

In this work, we primarily employ LLMs as embedding models and also consider two classical language models including GloVe and BERT. Detailed information about the embedding models used in this work is summarized in Table~\ref{tab-embedding}. Building upon Llama-2-7B-chat~\cite{touvron2023llama}, Mistral-7B-Instruct-v0.2~\cite{jiang2023mistral7b} and Llama-3-8B-Instruct~\cite{llama3modelcard}, we employ their fine-tuned versions tailored for text embedding from LLM2Vec~\footnote{https://github.com/McGill-NLP/llm2vec/tree/main}~\cite{llm2vec}.Additionally, three specialized text embedding models~\footnote{https://platform.openai.com/docs/guides/embeddings} (text-embedding-3-small, text-embedding-ada-002, text-embedding-3-large) from OpenAI are also evaluated in this work. For brevity, we refer to these models as ``small, ada, large'' in the following sections. Our text representation pipeline involves two sequential stages: (1) extracting token-level embeddings via the embedding models, followed by (2) applying pooling strategies to aggregate the token embeddings into final text representations. As detailed in Table~\ref{tab-embedding}, three distinct pooling strategies are evaluated in this benchmark. For each AD dataset, we derive 33 distinct text representations, each generated by varying combinations of embedding models and pooling strategies. Notably, the terms ``mntp''\footnote{mntp: masked next token prediction}, ``mntp-unsup-simcse''\footnote{mntp-unsup-simcse: unsupervised contrastive learning} and ``mntp-supervised''\footnote{mntp-supervised: supervised contrastive learning} denote the different fine-tuning techniques. More details on these fine-tuning techniques can be found in LLM2Vec~\cite{llm2vec}.

\begin{table}[h!]
    \centering
    \caption{Basic information of embedding models.}
    \label{tab-embedding}
    \resizebox{\textwidth}{!}{
    \begin{tabular}{c|ccccc}
    \toprule
    \textbf{Embedding Models} & \textbf{Pooling} &  \textbf{Max Tokens} & \textbf{Dimensions} & \textbf{Parameters} & \textbf{Open Source} \\
    \midrule
    GloVe (glove-6B) & Mean & 512 & 300 & 120M & \checkmark \\
    BERT (large-uncased) & Mean, CLS & 512 & 1024 & 340M & \checkmark \\
    LLaMA-2(mntp) & Mean, Weighted Mean, EOS & 8191 & 4096 & 7B & \checkmark \\
    LLaMA-2(mntp-unsup-simcse) & Mean, Weighted Mean, EOS & 8191 & 4096 & 7B & \checkmark \\
    LLaMA-2(mntp-supervised) & Mean, Weighted Mean, EOS & 8191 & 4096 & 7B & \checkmark \\
    LLaMA-3(mntp) & Mean, Weighted Mean, EOS & 8191 & 4096 & 8B & \checkmark \\
    LLaMA-3(mntp-unsup-simcse) & Mean, Weighted Mean, EOS & 8191 & 4096 & 8B & \checkmark \\
    LLaMA-3(mntp-supervised) & Mean, Weighted Mean, EOS & 8191 & 4096 & 8B & \checkmark \\
    Mistral(mntp) & Mean, Weighted Mean, EOS & 32768 & 4096 & 7B & \checkmark \\
    Mistral(mntp-unsup-simcse) & Mean, Weighted Mean, EOS & 32768 & 4096 & 7B & \checkmark \\
    Mistral(mntp-supervised) & Mean, Weighted Mean, EOS & 32768 & 4096 & 7B & \checkmark \\
    OpenAI (embedding-3-small) & - & 8192 & 1536 & - & \usym{2613} \\
    OpenAI (embedding-ada-002) & - & 8192 & 1536 & - & \usym{2613} \\
    OpenAI (embedding-3-large) & - & 8192 & 3072 & - & \usym{2613} \\
    \bottomrule
    \end{tabular}
    }
\end{table}

\subsection{Anomaly Detection Methods}
For comprehensive benchmarking, we consider both shallow machine learning-based AD algorithms, deep learning-based AD methods, and specialized textual AD methods. The details are described below.

\paragraph{Shallow Machine Learning-based AD Algorithms:}
    
\begin{itemize}

    \item \textbf{One-Class SVM (OCSVM)}~\cite{scholkopf1999support} OCSVM maps input data to a high-dimensional space by kernel methods and finds a hyperplane that separate normal data from the origin with maximum margin. RBF kernel was used in our experiments.
    
    \item \textbf{Isolation Forest (IForest)}~\cite{liu2008isolation} IForest isolates instances by building an ensemble of trees, then anomalies are those instances which have short average path lengths on the Trees. 

    \item \textbf{Local Outlier Factor (LOF)}~\cite{breunig2000lof} LOF measures the local deviation of the density of each instance with respect to its neighbors. We set the hyperparameter ``n\_neighbors''=30 in our experiments. 

    \item \textbf{Principal Component Analysis (PCA)}~\cite{shyu2003novel} PCA is a linear dimensionality reduction method that employs matrix decomposition to transform data into a new orthogonal coordinate system defined by its principal components. These principal components are ordered such that the first component captures the maximal variance in the data, with subsequent components representing decreasingly smaller variations. When applied in anomaly detection, PCA projects data onto a lower-dimensional subspace spanned by the top-k eigenvectors. The anomaly score for a new sample is computed as the reconstruction error.
    
    \item \textbf{K-Nearest Neighbors (KNN)}~\cite{ramaswamy2000efficient} KNN views the anomaly score of a new sample as the distance to its $k$-th nearest neighbor. We set $k=3$ in our experiments. 

    \item \textbf{Kernel Density Estimation (KDE)}~\cite{kim2012robust} KDE is a non-parametric method to estimate the probability density function (PDF) of a dataset. It smooths observed data points using a kernel function to approximate the underlying distribution. When applied in anomaly detection, anomalies are identified as points in low-density regions of the estimated PDF. We use the Gaussian kernel in our experiments.

    \item \textbf{Empirical-Cumulative-distributed-based Outlier Detection (ECOD)}~\cite{li2022ecod} ECOD is a hyperparameter-free outlier detection algorithm based on the empirical cumulative distribution function (ECDF). It employs ECDF to estimate the density of each feature independently and assumes that outliers are located in the tails of the distribution.
\end{itemize}

\paragraph{Deep Learning-based AD Methods:}

\begin{itemize}
    \item \textbf{AutoEncoder (AE)}~\cite{hinton2006reducing} An AutoEncoder is a type of neural network that consists of two parts including an encoder and a decoder. Encoder learns to compress input data into a lower-dimensional representation space and then decoder reconstructs them back to the original data space. When applied in anomaly detection, reconstruction error is a natural choice as the anomaly score. 

    \item \textbf{Deep Support Vector Data Description (SVDD)}~\cite{ruff2018deep} Deep SVDD is a deep one-class learning method that learns a minimal enclosing hypersphere in a latent space to characterize normal data. The anomaly score for a sample is given by the squared Euclidean distance to the hypersphere center in latent space.

    \item \textbf{Dense Projection for Anomaly Detection (DPAD)}~\cite{fu2024dense} DPAD is a density based method that learns to obtain locally dense low-dimensional representation for training data (normal data). The anomaly score is the KNN score in the low-dimensional space.
\end{itemize}

\paragraph{Specialized Textual AD Methods:}
\begin{itemize}

    \item \textbf{Context Vector Data Description (CVDD)}~\cite{ruff2019self} CVDD is a method that builds upon pretrained word embeddings to learn multiple sentence representations (context vector) that capture multiple semantic contexts via the self-attention mechanism and to project the word representations near these context vector by minimizing the cosine distance between them.

    \item \textbf{Detecting Anomalies in Text using ELECTRA (DATE)}~\cite{manolache2021date} DATE an end-to-end approach for the discrete text domain that combines a powerful representation learner for Text (ELECTRA)~\cite{clark2020electra} and an anomaly score is tailored for sequential data.

\end{itemize}

To evaluate these anomaly detection approaches, for seven shallow and three deep learning-based AD methods, we combine them with all 33 text embeddings, which results in 330 distinct two-stage anomaly detection configurations (10 AD methods $\times$ 33 embeddings) per dataset. Regarding the specialized text AD methods, CVDD utilizes word embeddings from GloVe and BERT, and DATE operates directly on raw text inputs without requiring pretrained embeddings.

\subsection{Evaluation Metrics}
For a holistic evaluation, this work utilizes both AUROC (Area Under the Receiver Operating Characteristic curve) and AUPRC (Area Under the Precision-Recall curve) to measure detection performance. 
AUROC is more suitable for scenarios where the number of positive and negative samples is relatively balanced or where high classification accuracy is required for both types of samples. It can comprehensively reflect the performance of the model. AUPRC is more suitable for scenarios with imbalanced positive and negative samples, especially where high precision and recall are required for the identification of positive samples. It can better reflect the model's predictive ability for the minority class.
The experiments on DBpedia14 were conducted on Intel(R) Xeon(R) Gold 5117 CPU @ 2.00GHz with 4$\times$ NVIDIA GeForce RTX 4090 GPUs, Python 3.8, and all other experiments all were conducted Intel(R) Xeon(R) CPU E5-2678 v3 @ 2.50GHz with 4$\times$ NVIDIA GeForce RTX 3090 GPUs, Python 3.8. We report the average results over five runs.

\section{Experimental results}

\subsection{Embedding-based text anomaly detection}
\begin{table}[h!]
    \centering
    \caption{Average AUROC(\%) of all two-stage methods. The top two results on each dataset are marked in \textbf{bold} (\textcolor{red!70}{top}, \textcolor{orange!70}{second}).}
    \label{tab-results-all-embeddings-auroc}
    \resizebox{\textwidth}{!}{
    \begin{tabular}{c|c|c|c|c|c|c|c|c|c|c|c|c|c}
    \toprule
    \textbf{Embedding} & \textbf{Pooling} & 20News. & Reuters. & IMDB & SST2 & SMS & Enron & WOS & DBp.0 & DBp.1 & DBp.2 & DBp.3 & DBp.4\\
    \midrule
    GloVe & Mean & 56.26 & 76.26 & 46.97 & 48.95 & 40.26 & 65.22 & 45.94 & 76.89 & 87.70 & 81.03 & 85.07 & 82.46 \\
    \midrule
    \multirow{2}{*}{BERT} & CLS & 55.73 & 60.05 & 45.27 & 53.99 & 41.84 & 60.11 & 47.62 & 85.41 & 85.96 & 77.30 & 74.33 & 74.19 \\
    & Mean & 54.89 & 80.33 & 45.40 & 54.25 & 51.77 & 59.88 & 48.19 & 83.41 & 85.92 & 83.66 & 82.48 & 82.00\\
    \midrule
    \multirow{3}{*}{\makecell{LLaMA-2\\(mntp)}} & EOS & 47.40 & 64.82 & 48.90 & 53.26 & 62.18 & 57.73 & 48.49 & 66.25 & 72.31 &  65.09 & 72.45 & 69.68\\
    & Mean & 51.95 & 61.16 & 48.03 & 52.29 & 56.07 & 60.36 & 47.15 & 66.95  & 74.54 & 72.93 & 80.45 & 80.20\\
    & Weighted Mean & 51.15 & 61.10 & 48.21 & 52.18 & 60.52 & 59.18 & 46.61 & 65.45 & 73.26 & 71.01 & 78.46 & 78.25\\
    \midrule
    \multirow{3}{*}{\makecell{LLaMA-2\\(mntp-unsup-simcse)}} & EOS & 59.30 & \cellcolor{red!30}\textbf{92.49} & 48.19 & 64.56 & 79.80 & 82.89 & 59.22 & \cellcolor{red!30}\textbf{92.66} & \cellcolor{orange!30}\textbf{98.00} & 88.81 & \cellcolor{red!30}\textbf{97.01} & 94.72\\    
    & Mean & 52.26 & 86.72 & 49.41 & 62.91 & 78.64 & 74.76 & 44.81 & 78.22 & 91.18 & 81.15 & 94.99 & 91.85\\
    & Weighted Mean & 53.02 & 85.27 & 49.93 & 63.11 & 84.83 & 72.49 & 44.87 & 77.34 & 90.86 & 79.05 & 94.11 & 90.97\\
    \midrule
    \multirow{3}{*}{\makecell{LLaMA-2\\(mntp-supervised)}} & EOS & 57.53 & 90.41 & 51.74 & 68.61 & \cellcolor{red!30}\textbf{91.40} & 82.74 & \cellcolor{red!30}\textbf{67.18} & 88.16 & 97.59 & 84.21 & 94.88 & 95.22\\
    & Mean & 56.81 & 89.84 & 53.38 & \cellcolor{orange!30}\textbf{71.35} & 78.93 & 80.53 & 49.95 & 83.96 & 97.20 & 87.67 & 93.44 & 93.50\\
    & Weighted Mean & 57.13 & 88.55 & 53.13 & 69.40 & 84.67 & 80.11 & 50.95 & 82.39 & 96.49 & 86.69 & 92.69 & 92.91\\
    \midrule
     \multirow{3}{*}{\makecell{LLaMA-3\\(mntp)}} & EOS & 62.57 & 82.40 & 45.53 & 56.90 & \cellcolor{orange!30}\textbf{90.86} & 74.07 & 55.80 & 76.92 & 89.09 & 87.06 & \cellcolor{orange!30}\textbf{95.20} & 90.57\\
    & Mean & 56.82 & 85.16 & 47.40 & 58.54 & 72.42 & 70.57 & 45.89 & 66.82 & 82.35 & 83.03 & 93.84 & 89.98\\
    & Weighted Mean & 54.59 & 81.82 & 47.72 & 59.11 & 80.20 & 67.15 & 46.30 & 67.44 & 82.70 & 81.68 & 93.16 & 90.05\\
    \midrule
    \multirow{3}{*}{\makecell{LLaMA-3\\(mntp-unsup-simcse)}} & EOS & 54.35 & 77.95 & 47.36 & 58.47 & 89.98 & 81.19 & 55.76 & 63.48 & 78.27 & 64.78 & 75.43 & 72.86\\
    & Mean & 48.68 & 86.53 & 49.26 & 60.56 & 78.63 & 72.96 & 42.38 & 64.76 & 86.10 & 80.40 & 94.79 & 91.02\\
    & Weighted Mean & 46.56 & 84.57 & 49.42 & 60.17 & 86.78 & 70.13 & 43.35 & 66.28 & 85.92 & 79.62 & 94.03 & 90.56\\
    \midrule
    \multirow{3}{*}{\makecell{LLaMA-3\\(mntp-supervised)}} & EOS & 62.17 & 89.04 & 52.98 & 67.22 & 90.70 & 80.01 & 56.24 & 87.97 & 97.31 & 82.84 & 91.48 & 92.61\\
    & Mean & 60.06 & 92.21 & 50.63 & 69.56 & 86.64 & 79.87 & \cellcolor{orange!30}\textbf{63.59} & 87.30 & \cellcolor{red!30}\textbf{98.16} & \cellcolor{red!30}\textbf{90.85} & 95.11 & \cellcolor{red!30}\textbf{96.04}\\
    & Weighted Mean & 57.91 & 90.95 & 50.58 & 68.84 & 86.28 & 79.98 & 62.76 & 87.37 & 97.88 & \cellcolor{orange!30}\textbf{89.90} & 94.57 & \cellcolor{orange!30}\textbf{95.42}\\
    \midrule
     \multirow{3}{*}{\makecell{Mistral\\(mntp)}} & EOS & \cellcolor{red!30}\textbf{71.99} & 76.96 & 43.67 & 58.10 & 87.77 & 80.89 & 59.08 & 79.32 & 90.56 & 86.33 & 91.47 & 88.10\\
    & Mean & \cellcolor{orange!30}\textbf{64.55} & 74.79 & 50.63 & 55.39 & 78.32 & 70.94 & 46.56 & 69.48 & 83.72 & 81.16 & 92.50 & 88.31\\
    & Weighted Mean & 62.19 & 71.77 & 50.60 & 54.86 & 80.42 & 68.24 & 46.53 & 67.95 & 83.10 & 80.12 & 91.79 & 88.33\\
    \midrule
    \multirow{3}{*}{\makecell{Mistral\\(mntp-unsup-simcse)}} & EOS & 56.89 & 85.30 & 52.79 & 64.78 & 84.88 & \cellcolor{red!30}\textbf{89.53} & 63.80 & 79.68 & 94.45 & 75.24 & 86.40 & 82.07\\
    & Mean & 51.04 & 78.63 & 53.92 & 64.58 & 76.65 & 75.98 & 39.71 & 65.20 & 81.42 & 72.29 & 87.38 & 82.09\\
    & Weighted Mean & 50.97 & 76.25 & 54.13 & 64.70 & 84.21 & 73.95 & 40.38 & 63.85 & 80.40 & 69.78 & 85.89 & 81.87\\
    \midrule
    \multirow{3}{*}{\makecell{Mistral\\(mntp-supervised)}} & EOS & 53.07 & 88.51 & \cellcolor{red!30}\textbf{56.46} & 71.12 & 83.70 & 81.25 & 60.27 & 87.57 & 95.57 & 88.31 & 94.46 & 91.52\\
    & Mean & 50.88 & 91.10 & 50.67 & \cellcolor{red!30}\textbf{71.98} & 65.36 & \cellcolor{orange!30}\textbf{84.37} & 52.86 & 83.67 & 94.65 & 87.77 & 92.47 & 90.17\\
    & Weighted Mean & 50.75 & 89.24 & 51.07 & 71.17 & 70.84 & 83.99 & 52.75 & 81.09 & 94.41 & 87.15 & 91.68 & 88.42\\
    \midrule
    OpenAI-3-small & - & 55.61 & 91.10 & 51.69 & 66.50 & 57.88 & 74.08 & 58.52 & 91.20 & 94.34 & 88.10 & 93.08 & 91.64\\
    OpenAI-ada-002 & - & 61.51 & 88.66 & 51.60 & 65.49 & 77.54 & 78.12 & 60.28 & 86.05 & 92.45 & 87.58 & 93.05 & 92.54\\
    OpenAI-3-large & - & 51.84 & \cellcolor{orange!30}\textbf{92.10} & \cellcolor{orange!30}\textbf{54.78} & 67.60 & 65.41 & 77.55 & 63.31 & \cellcolor{orange!30}\textbf{92.52} & 94.15 & 88.03 & 93.43 & 91.58\\
    \bottomrule
    \end{tabular}
    }
\end{table}

\begin{table}[h!]
    \centering
    \caption{Average AUPRC(\%) of all two-stage methods. The top two results on each dataset are marked in \textbf{bold} (\textcolor{red!70}{top}, \textcolor{orange!70}{second}).}
    \label{tab-results-all-embeddings-AUPRC}
    \resizebox{\textwidth}{!}{
    \begin{tabular}{c|c|c|c|c|c|c|c|c|c|c|c|c|c}
    \toprule
    \textbf{Embedding} & \textbf{Pooling} & 20News. & Reuters. & IMDB & SST2 & SMS & Enron & WOS & DBp.0 & DBp.1 & DBp.2 & DBp.3 & DBp.4\\
    \midrule
    GloVe & Mean & 13.62 & 83.03 & 59.97 & 51.77 & 26.40 & 39.46 & 29.55 & 71.45 & 82.95 & 75.84 & 81.68 & 77.87\\
    \midrule
    \multirow{2}{*}{BERT} & CLS & 14.51 & 66.85 & 59.59 &  54.69 & 25.07 & 38.22 & 30.71 & 82.73 & 81.00 & 72.35 & 68.03 & 66.32\\
    & Mean & 14.05 & 83.73 & 58.99 & 54.50 & 31.27 & 37.28 & 32.26 & 78.99 & 80.30 & 79.26 & 77.77 & 74.87\\
    \midrule
    \multirow{3}{*}{\makecell{LLaMA-2\\(mntp)}} & EOS & 10.99 & 75.29 & 61.57 & 54.04 & 38.91 & 39.29 & 29.66 & 56.95 & 64.90 & 58.91 & 67.42 & 63.20 \\
    & Mean & 12.31 & 72.68 & 60.86 & 53.47 & 38.03 & 37.47 & 28.99 & 56.74 & 66.76 & 65.35 & 77.06 & 73.65\\
    & Weighted Mean & 12.15 & 72.37 & 61.02 & 53.25 & 43.37 & 37.19 & 28.65 & 56.02 & 66.11 & 63.95 & 75.18 & 71.15\\
    \midrule
    \multirow{3}{*}{\makecell{LLaMA-2\\(mntp-unsup-simcse)}} & EOS & 14.69 & 93.86 & 60.98 & 62.97 & 57.42 & 64.46 & 38.03 & 90.79 & \cellcolor{orange!30}\textbf{97.47} & 86.86 & \cellcolor{red!30}\textbf{96.33} & 92.88\\    
    & Mean & 12.78 & 88.45 & 61.47 & 62.98 & 59.27 & 52.26 & 28.30 & 71.08 & 88.06 & 77.33 & 93.84 & 88.99\\
    & Weighted Mean & 13.01 & 87.46 & 61.76 & 62.86 & 68.98 & 51.36 & 28.27 & 70.44 & 88.16 & 74.64 & 92.81 & 87.80\\
    \midrule
    \multirow{3}{*}{\makecell{LLaMA-2\\(mntp-supervised)}} & EOS & 16.44 & 93.34 & 63.17 & 66.15 & 77.74 & 64.81 & \cellcolor{orange!30}\textbf{45.06} & 84.99 & 96.63 & 83.42 & 94.39 & 93.76\\
    & Mean & 16.00 & 93.05 & 64.39 & \cellcolor{orange!30}\textbf{69.89} & 57.70 & 59.51 & 32.04 & 80.70 & 96.40 & 85.42 & 92.26 & 92.09\\
    & Weighted Mean & 16.67 & 92.28 & 64.26 & 67.85 & 66.43 & 59.85 & 32.63 & 79.16 & 95.58 & 84.22 & 91.06 & 91.29 \\
    \midrule
     \multirow{3}{*}{\makecell{LLaMA-3\\(mntp)}} & EOS & 18.43 & 85.76 & 59.26 & 57.55 & \cellcolor{orange!30}\textbf{78.56} & 56.45 & 35.15 & 67.60 & 82.73 & 82.22 & 92.94 & 86.25\\
    & Mean & 14.53 & 87.69 & 60.23 & 59.10 & 46.72 & 47.12 & 29.02 & 56.89 & 75.39 & 77.99 & 92.22 & 85.68\\
    & Weighted Mean & 13.97 & 85.18 & 60.37 & 59.30 & 56.77 & 45.10 & 28.88 & 57.53 & 76.42 & 76.86 & 91.41 & 85.93\\
    \midrule
    \multirow{3}{*}{\makecell{LLaMA-3\\(mntp-unsup-simcse)}} & EOS & 13.28 & 82.85 & 60.27 & 57.92 & 77.25 & 66.06 & 36.67 & 57.35 & 71.57 & 58.27 & 68.15 & 65.09\\
    & Mean & 12.07 & 88.82 & 61.28 & 60.82 &  58.63 & 51.45 & 27.78 & 57.52 & 82.19 & 78.73 & 93.93 & 89.08\\
    & Weighted Mean & 11.53 & 87.49 & 61.26 & 60.21 & 70.42 & 50.34 & 27.96 & 58.80 & 82.46 & 76.84 & 92.98 & 88.22\\
    \midrule
    \multirow{3}{*}{\makecell{LLaMA-3\\(mntp-supervised)}} & EOS & 18.38 & 93.08 & 64.30 & 66.75 & \cellcolor{red!30}\textbf{81.73} & 63.40 & 35.55 & 86.77 & 96.76 & 77.69 & 88.92 & 90.01\\
    & Mean & 18.90 & 94.93 & 63.07 & 68.39 & 75.57 & 63.20 &   42.54 & 84.23 & \cellcolor{red!30}\textbf{97.62} & \cellcolor{red!30}\textbf{89.40} & 94.20 & \cellcolor{red!30}\textbf{95.11}\\
    & Weighted Mean & 17.24 & 94.09 & 63.04 & 67.86 & 74.50 & 63.32 &  41.61 & 84.61 & 97.37 & 88.10 & 93.43 & \cellcolor{orange!30}\textbf{94.43}\\
    \midrule
     \multirow{3}{*}{\makecell{Mistral\\(mntp)}} & EOS & \cellcolor{red!30}\textbf{23.38} & 84.77 & 57.89 & 57.71 & 72.76 & 65.53 & 38.72 & 73.90 & 86.92 & 84.52 & 90.61 & 85.90\\
    & Mean & 16.84 & 79.46 & 62.53 & 55.88 & 54.97 & 46.45 & 29.20 & 59.40 & 74.81 & 74.21 & 89.55 & 81.57\\
    & Weighted Mean & 16.08 & 77.65 & 62.55 & 55.52 & 58.02 & 44.84 & 29.25 & 57.82 & 74.88 & 72.74 & 88.81 & 81.92\\
    \midrule
    \multirow{3}{*}{\makecell{Mistral\\(mntp-unsup-simcse)}} & EOS & 15.65 & 89.32 & 63.84 & 62.75 & 68.35 & \cellcolor{orange!30}\textbf{78.10} &  41.85 & 74.75 & 91.78 & 72.03 & 84.83 & 79.31\\
    & Mean & 12.37 & 83.75 & 64.48 & 63.99 & 57.76 &  57.44 &  26.60 & 57.14 & 76.79 & 66.83 & 84.69 & 78.56\\
    & Weighted Mean & 12.32 & 82.29 & 64.60 & 63.76 & 69.51 & 56.85 &  26.66 & 56.55 & 76.26 & 63.28 & 82.76 & 77.12\\
    \midrule
    \multirow{3}{*}{\makecell{Mistral\\(mntp-supervised)}} & EOS & 14.20 & 92.42 & \cellcolor{red!30}\textbf{67.12} & 68.78 & 67.82 & \cellcolor{orange!30}\textbf{67.41} & 38.98 & 84.36 & 94.50 & 86.62 & 93.57 & 90.07\\
    & Mean & 12.55 & 94.15 & 62.69 & \cellcolor{red!30}\textbf{70.39} & 42.49 & 66.87 & 33.17 & 79.29 & 93.36 & 84.55 & 90.88 & 88.17\\
    & Weighted Mean & 12.53 & 92.84 & 62.93 & 69.54 & 48.78 & 66.84 & 33.05 & 77.46 & 93.25 & 83.42 & 89.67 & 86.33\\
    \midrule
    OpenAI-3-small & - & 19.04 & \cellcolor{orange!30}\textbf{95.35} & 65.29 & 67.93 & 38.38 & 52.64 & 40.09 & \cellcolor{red!30}\textbf{92.24} & 96.27 & \cellcolor{orange!30}\textbf{89.35} & 94.94 & 92.76\\
    OpenAI-ada-002 & - & \cellcolor{orange!30}\textbf{21.83} & 93.08 & 64.43 & 66.24 & 54.38 & 59.43 & 42.37 & 83.67 & 92.01 & 88.47 & \cellcolor{orange!30}\textbf{94.99} & 93.16\\
    OpenAI-3-large & - & 17.58 & \cellcolor{red!30}\textbf{96.10} & \cellcolor{orange!30}\textbf{66.96} & 69.12 & 42.00 & 56.88 & \cellcolor{red!30}\textbf{45.75} & \cellcolor{orange!30}\textbf{91.64} & 93.02 & 83.55 & 93.88 & 91.17\\
    \bottomrule
    \end{tabular}
    }
\end{table}

Table~\ref{tab-results-all-embeddings-auroc} and Table~\ref{tab-results-all-embeddings-AUPRC} report average AUROC (\%) and AUPRC (\%) on all two-stage detectors across different embeddings and datasets, respectively. Depending on the results of Table~\ref{tab-results-all-embeddings-auroc} and Table~\ref{tab-results-all-embeddings-AUPRC}, we have the following observations and analysis:

\begin{itemize}
    \item Across all datasets, the top two performing results originate from the detectors utilizing LLM-derived embeddings. This indicates that LLM-based embeddings boost the performance for two-stage anomaly detection frameworks and exhibit marked advantages over traditional embedding methods such as GloVe and BERT for text anomaly detection tasks.

    \item No single LLM-derived embedding universally outperforms others across all datasets. This suggests that the optimal choice of embedding model may depend on specific dataset characteristics or task requirements.

    \item The results of AUPRC (\%) in Table~\ref{tab-results-all-embeddings-AUPRC} reveal that the embedding models from OpenAI perform much better than other embedding models, which suggests that the models from OpenAI achieve both high precision and high recall, a critical advantage in the scenarios where false negatives are costly. 
 \end{itemize}


\begin{figure}[h!]
    \centering
    \includegraphics[width=0.95\linewidth]{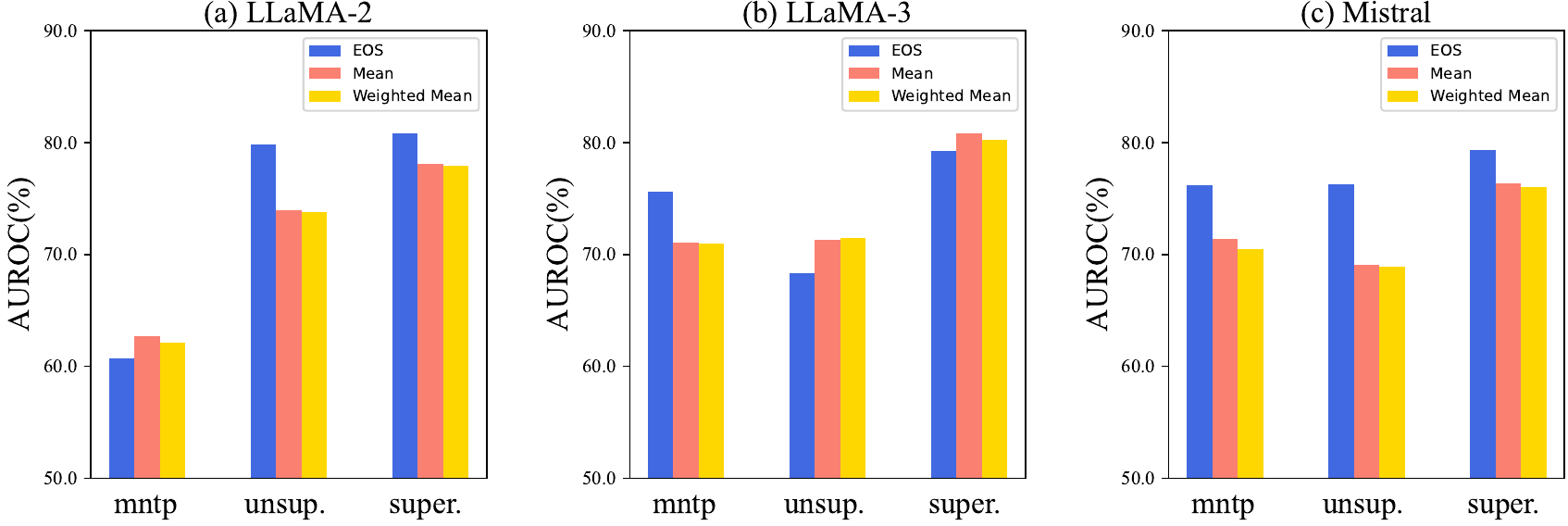}
    \caption{The average AUROC(\%) across all datasets. The ``unsup.'' and ``super.'' correspond to the embedding models with ``mntp-unsup-simcse'' and ``mntp-supervised'', respectively.}
    \label{fig-mean-pooling}
\end{figure}

\begin{figure}[h!]
    \centering
    \includegraphics[width=\linewidth]{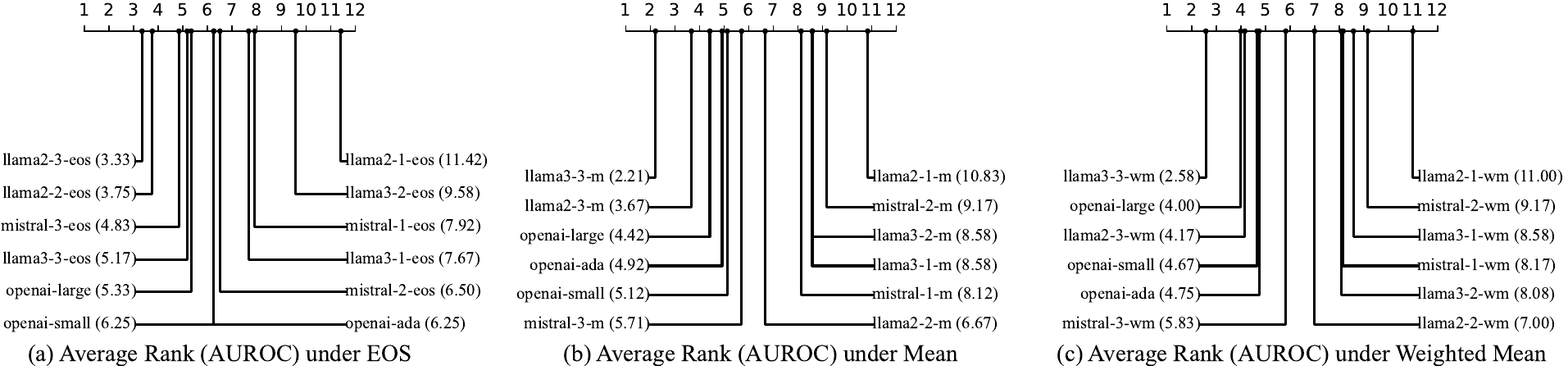}
    \caption{Performance comparison of different embeddings from LLMs. Note that ``llama2-1, llama3-1, mistral-1'' denote those corresponding embedding models with ``mntp'', and ``llama2-2, llama3-2, mistral-2'' denote those corresponding embedding models with ``mntp-unsup-simcse'' and ``llama2-3, llama3-3, mistral-3'' denote those corresponding embedding models with ``mntp-supervised''. The plots (a), (b), (c) compare the average AUROC from all the embedding models under EOS, Mean, and Weighted Mean, respectively.}
    \label{fig-rank-embedding}
\end{figure}

To observe the effect of different pooling strategies across all datasets, we compute the row-wise averages based on Table~\ref{tab-results-all-embeddings-auroc} to get the mean performance of each embedding-pooling combination across all datasets. We visualize these aggregated mean results for the LLaMA-2, LLaMA-3, and Mistral in Figure~\ref{fig-mean-pooling}, enabling direct comparison for the overall effectiveness of different pooling strategies, where ``EOS'' exhibits significant advantages over ``Mean'' and ``Weighted Mean'' in most cases. In addition, we also notice that ``Mean'' and ``Weighted Mean'' have quite similar performance in all cases. Figure~\ref{fig-rank-embedding} presents the average AUROC ranking across different LLM-derived embeddings. The visualization reveals that embeddings fine-tuned using the ``mntp-supervised'' approach (denoted as -3-) consistently achieve superior ranking positions compared to other embeddings. This finding aligns with the results reported in ~\cite{llm2vec}, where representations fine-tuned by ``mntp-supervised'' achieved state-of-the-art performance on the Massive Text Embedding Benchmark (MTEB).

\begin{figure}[h!]
    \centering
    \includegraphics[width=0.95\linewidth]{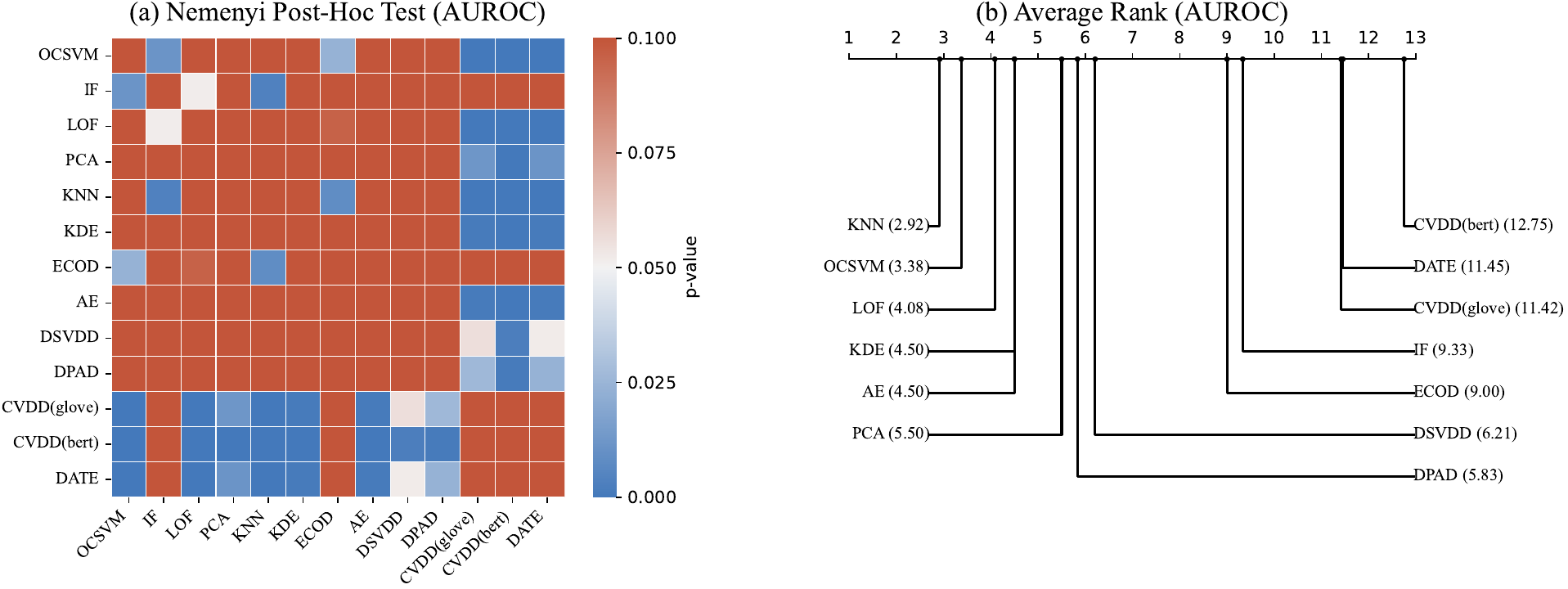}
    \caption{Performance comparison of different AD methods. In plot (a), a pair of methods with p-value > 0.05 indicates no statistically significant difference in their detection performance at the 95\% confidence level.}
    \label{fig-nemenyi-rank}
\end{figure} 

\begin{table}[h!]
    \centering
    \caption{Comparison among different AD methods. The top two results for each dataset are marked in \textbf{bold} (\textcolor{red!70}{top}, \textcolor{orange!70}{second}). Note that ``mean'' denotes the average performance on all embeddings and ``best'' refers to the best performance across all embeddings.}
    \label{tab-results-all-methods}
    \resizebox{\textwidth}{!}{
    \begin{tabular}{l|c|c|c|c|c|c|c|c|c|c|c|c|c}
    \toprule
    \textbf{Methods} & 20News. & Reuters. & IMDB & SST2 & SMS & Enron & WOS & DBp.0 & DBp.1 & DBp.2 & DBp.3 & DBp.4 & Avg\\
    \midrule
    OCSVM (GloVe) & 55.88 & 76.85 & 46.35 & 46.38 & 48.02 & 63.12 & 52.49 & 82.33 & 91.59 & 85.27 & 87.97 & 86.64 & 68.57\\
    OCSVM (BERT) & 52.80 & 80.74 & 45.11 & 51.94 & 54.87 & 53.26 & 53.09 & 81.63 & 82.06 & 80.26 & 75.88 & 76.02 & 65.64\\
    OCSVM (mean) & 55.69 & 85.33 & 49.47 & 63.04 & 81.94 & 73.90 & 51.52 & 79.87 & 90.66 & 83.74 & 91.14 & 89.98 & 74.69\\
    OCSVM (best) & 72.74 & \cellcolor{red!30}\textbf{99.22} & 62.97 & \cellcolor{red!30}\textbf{85.22} & 94.66 & \cellcolor{red!30}\textbf{99.24} & 69.13 & 98.70 & 99.87 & 95.83 & 99.46 & 98.89 & \cellcolor{orange!30}\textbf{89.66}\\
    
    \midrule
    IForest (GloVe) & 56.09 & 78.21 & 46.76 & 44.69 & 33.43 & 62.45 & 38.61 & 77.03 & 88.37 & 83.18 & 82.11 & 82.12 & 64.42\\
    IForest (BERT) & 55.17 & 80.79 & 45.34 & 50.73 & 46.89 & 54.83 & 50.35 & 78.96 & 77.26 & 76.38 & 72.88 & 72.27 & 63.49\\
    IForest (mean) & 54.42 & 80.39 & 48.20 & 55.86 & 72.32 & 64.73 & 48.01 & 70.05 & 83.63 & 73.82 & 82.84 & 81.24 & 67.96\\
    IForest (best) & 71.87 & 94.08 & 54.68 & 66.90 & 92.83 & 75.01 & 62.69 & 84.70 & 95.89 & 84.84 & 93.48 & 89.15 & 80.51\\
    
    \midrule
    LOF (GloVe) & 60.16 & 84.54 & 49.31 & 58.73 & 44.65 & 67.86 & 64.93 & 86.25 & 96.57 & 87.29 & 97.15 & 92.22 & 74.14\\
    LOF (BERT) & 58.86 & 89.45 & 47.75 & 58.66 & 56.05 & 76.18 & 59.02 & 91.35 & 95.04 & 92.97 & 96.42 & 93.65 & 76.28\\
    LOF (mean) & 54.87 & 80.66 & 54.30 & 64.83 & 77.35 & 81.31 & 62.12 & 87.77 & 94.73 & 90.42 & 96.73 & 94.26 & 78.28\\
    LOF (best) & 70.79 & 93.86 & \cellcolor{orange!30}\textbf{64.66} & 71.96 & 93.62 & 94.52 & 73.27 & \cellcolor{red!30}\textbf{99.30} & \cellcolor{orange!30}\textbf{99.89} & \cellcolor{red!30}\textbf{97.91} & \cellcolor{red!30}\textbf{99.81} & 99.22 & 88.23\\
    \midrule
    PCA (GloVe) & 56.36 & 77.77 & 46.51 & 44.79 & 34.60 & 62.64 & 39.30 & 78.19 & 90.11 & 83.39 & 83.69 & 83.09 & 65.04\\
    PCA (BERT) & 54.21 & 80.45 & 44.59 & 51.45 & 50.44 & 55.01 & 49.05 & 80.39 & 80.88 & 78.62 & 70.94 & 72.75 & 64.06\\
    PCA (mean) & 55.66 & 83.99 & 48.02 & 57.57 & 78.04 & 65.66 & 48.11 & 77.04 & 88.63 & 80.97 & 88.73 & 87.43 & 71.65\\
    PCA (best) & 73.68 & 99.13 & 56.58 & 69.76 & 95.01 & 79.23 & 67.61 & 98.50 & 99.83 & 94.18 & 99.02 & 98.42 & 85.91\\
    
    \midrule
    KNN (GloVe) & 59.90 & 82.75 & 46.74 & 48.68 & 35.43 & 73.63 & 33.74 & 79.83 & 89.22 & 83.86 & 92.02 & 85.92 & 67.64\\
    KNN (BERT) & 55.61 & 83.61 & 47.65 & 59.00 & 33.36 & 61.86 & 37.77 & 87.41 & 92.76 & 89.61 & 95.96 & 93.23 & 69.82\\
    KNN (mean) & 59.40 & 85.34 & 56.82 & 68.89 & 74.57 & 78.89 & 56.09 & 84.32 & 93.50 & 86.97 & 95.10 & 93.02 & 77.74\\
    KNN (best) & \cellcolor{red!30}\textbf{74.56} & 96.58 & \cellcolor{red!30}\textbf{73.44} & 80.11 & 94.61 & 92.78 & \cellcolor{red!30}\textbf{80.91} & \cellcolor{orange!30}\textbf{99.19} & \cellcolor{red!30}\textbf{99.90} & \cellcolor{orange!30}\textbf{97.54} & 99.64 & \cellcolor{orange!30}\textbf{99.24} & \cellcolor{red!30}\textbf{90.71}\\
    
    \midrule
    KDE (GloVe) & 55.99 & 75.96 & 46.37 & 45.23 & 41.38 & 63.57 & 47.93 & 81.04 & 90.89 & 84.03 & 85.17 & 84.82 & 66.86\\
    KDE (BERT) & 53.90 & 78.06 & 44.29 & 51.32 & 51.41 & 54.88 & 49.52 & 80.11 & 79.44 & 77.28 & 69.22 & 71.38 & 63.40\\
    KDE (mean) & 56.04 & 84.09 & 48.84 & 62.06 & 81.43 & 71.38 & 48.67 & 78.68 & 89.24 & 81.94 & 89.55 & 88.47 & 73.37\\
    KDE (best) & 73.18 & 99.04 & 61.74 & \cellcolor{orange!30}\textbf{84.67} & \cellcolor{red!30}\textbf{95.38} & \cellcolor{orange!30}\textbf{99.11} & 68.70 & 98.59 & 99.84 & 94.03 & 98.99 & 98.40 & 89.31\\
    
    \midrule
    ECOD (GloVe) & 55.42 & 64.57 & 42.95 & 43.83 & 32.81 & 58.58 & 36.11 & 65.91 & 76.42 & 74.83 & 69.89 & 70.75 & 57.67\\
    ECOD (BERT) & 52.41 & 66.28 & 39.24 & 48.39 & 32.27 & 51.93 & 43.80 & 71.79 & 70.92 & 71.73 & 56.43 & 63.34 & 55.71\\
    ECOD (mean) & 53.05 & 71.35 & 40.10 & 51.79 & 59.89 & 61.85 & 43.24 & 64.61 & 79.13 & 72.60 & 79.69 & 79.91 & 63.10\\
    ECOD (best) & 71.97 & 93.74 & 48.10 & 57.75 & 89.95 & 75.51 & 63.07 & 91.73 & 99.57 & 87.13 & 96.97 & 95.86 & 80.95\\
    
    \midrule
    AE (GloVe) & 55.60 & 80.93 & 48.66 & 49.60 & 31.25 & 69.41 & 40.26 & 79.30 & 89.48 & 83.76 & 91.06 & 86.25 & 67.13\\
    AE (BERT) & 54.84 & 83.64 & 46.87 & 57.39 & 42.61 & 59.28 & 41.25 & 90.65 & 94.50 & 93.38 & 96.35 & 94.07 & 71.24\\
    AE (mean) & 57.28 & 87.62 & 53.86 & 71.12 & 75.08 & 87.45 & 56.29 & 85.17 & 94.24 & 88.99 & 95.77 & 93.68 & 78.88\\
    AE (best) & 71.51 & 98.75 & 59.17 & 80.95 & 94.37 & 98.72 & \cellcolor{orange!30}\textbf{75.89} & 98.32 & 99.75 & 97.14 & \cellcolor{orange!30}\textbf{99.71} & 99.15 & 89.45\\
    
    \midrule
    DSVDD (GloVe) & 49.77 & 57.92 & 47.72 & 53.47 & 64.28 & 56.91 & 56.32 & 56.61 & 68.34 & 57.31 & 66.49 & 61.66 & 58.07\\
    DSVDD (BERT) & 54.49 & 71.45 & 46.19 & 51.35 & 89.43 & 54.60 & 49.32 & 81.18 & 91.28 & 84.91 & 93.83 & 88.91 & 71.41\\
    DSVDD (mean) & 54.32 & 80.11 & 49.49 & 58.62 & 76.25 & 70.74 & 52.04 & 70.89 & 84.70 & 74.30 & 87.78 & 81.38 & 70.05\\
    DSVDD (best) & 68.32 & \cellcolor{orange!30}\textbf{99.12} & 54.06 & 67.80 & 92.98 & 85.81 & 59.92 & 98.27 & 99.87 & 95.31 & 99.69 & \cellcolor{red!30}\textbf{99.26} & 85.03\\
    
    \midrule
    DPAD (GloVe) & 57.43 & 83.08 & 48.31 & 54.09 & 36.78 & 73.99 & 49.69 & 82.42 & 96.02 & 87.38 & 95.19 & 91.17 & 71.30\\
    DPAD (BERT) & 56.64 & 88.87 & 46.96 & 62.26 & 60.38 & 79.74 & 48.70 & 90.67 & 92.02 & 91.41 & 96.92 & 94.38 & 75.75\\
    DPAD (mean) & 56.36 & 82.95 & 51.32 & 67.58 & 76.57 & 86.76 & 54.23 & 78.27 & 88.81 & 81.59 & 90.13 & 86.90 & 75.12\\
    DPAD (best) & \cellcolor{orange!30}\textbf{73.41} & 94.56 & 57.02 & 78.36 & \cellcolor{orange!30}\textbf{95.02} & 97.43 & 66.74 & 93.53 & 99.25 & 93.59 & 99.62 & 98.61 & 87.26\\
    
    \midrule
    CVDD (GloVe) & 62.60 & 86.48 & 46.66 & 51.71 & 49.00 & 69.38 & 50.51 & 91.26 & 97.13 & 74.83 & 92.03 & 82.14 & 71.14\\
    CVDD (BERT) & 54.98 & 51.76 & 47.64 & 46.83 & 44.80 & 52.86 & 40.23 & 57.24 & 61.44 & 56.21 & 55.69 & 52.47 & 51.85\\
    DATE & 51.47 & - & 48.56 & 56.96 & 72.89 & 71.67 & 56.36 & 79.61 & 85.92 & 77.18 & 84.14 & 78.31 & 69.37\\
    \bottomrule
    \end{tabular}
    }
\end{table}

\begin{figure}[h!]
    \centering
    \includegraphics[width=0.95\linewidth]{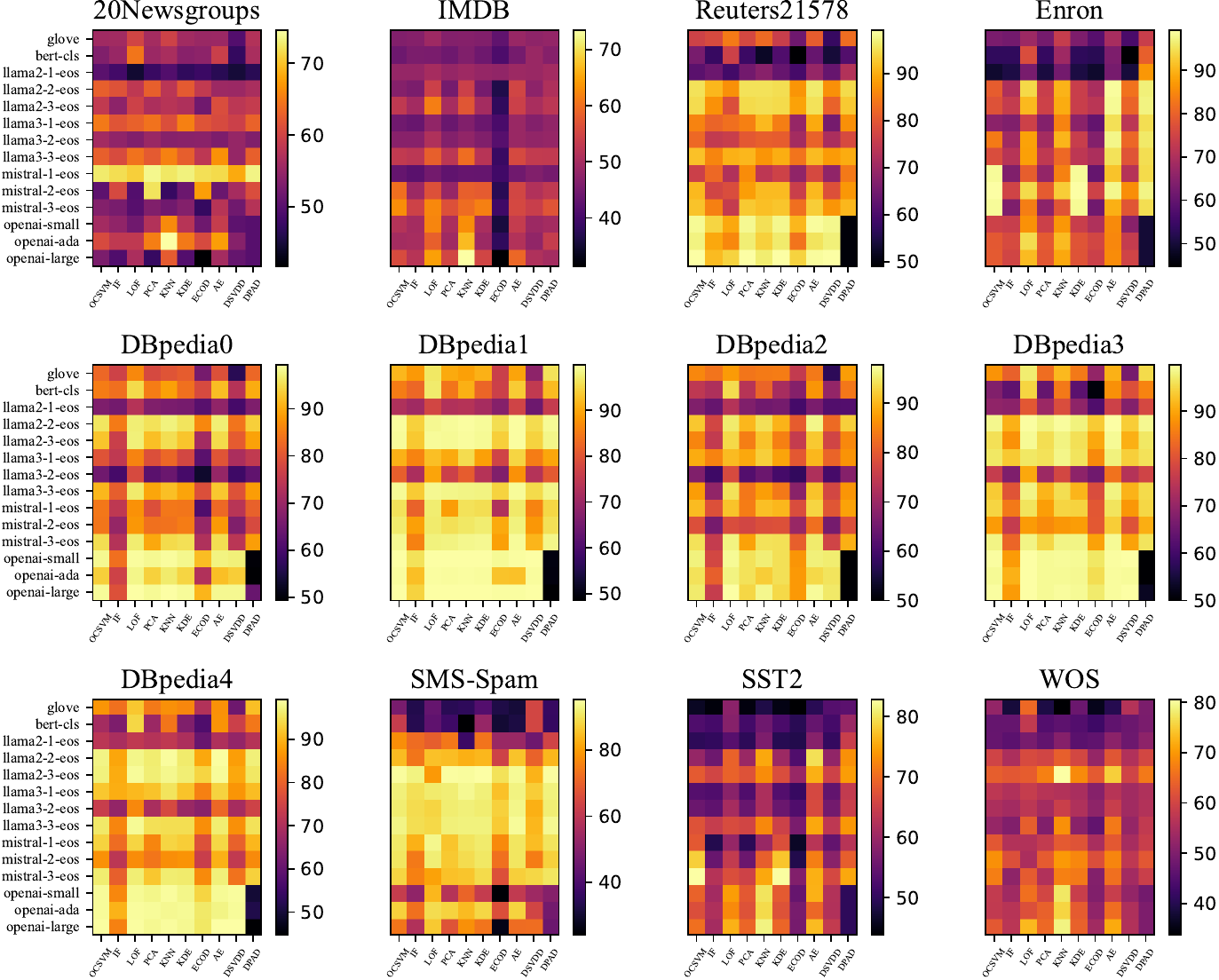}
    \caption{The heatmap of AUROC(\%) of each two-stage AD method with different text embeddings. Note that ``llama2-1, llama3-1, mistral-1'' denote those corresponding ``mntp'' models, ``llama2-2, llama3-2, mistral-2'' denote those corresponding ``mntp-unsup-simcse'' models and ``llama2-3, llama3-3, mistral-3'' denote those corresponding ``mntp-supervised'' models.}
    \label{fig-auroc-heatmap}
\end{figure}

Table~\ref{tab-results-all-methods} summarizes detection performance (AUROC\%) of different AD methods across all datasets. Notably, CVDD, a specialized text anomaly detection method, operates directly on token-level embedding rather than aggregated sentence-level representations. Due to computational constraints associated with processing all token embeddings from LLMs, the experiments on CVDD were limited to conventional embeddings, specifically GloVe and BERT. Futhermore, DATE, another text-specific AD method, requires access to raw text data rather than precomputed embeddings. 
Figure~\ref{fig-nemenyi-rank} presents a comprehensive comparison of AD methods based on the results from Table~\ref{tab-results-all-methods} and two-stage AD methods only use the ``best'' results. In Figure~\ref{fig-nemenyi-rank}, plot (a), a heatmap visualization of p-values derived from the Nemenyi post-hoc test, and plot (b), an average rank diagram where lower rank indicates better performance. Depending on the results of Table~\ref{tab-results-all-methods} and Figure~\ref{fig-nemenyi-rank}, we have the following observations: 

\begin{itemize}
    \item Analyzing the ``best'' results of all two-stage AD methods, deep learning-based detectors (AE, DSVDD, DPAD) exhibit no advantage over conventional ``shallow'' algorithms(OCSVM, IForest, LOF, KNN, KDE) when using LLM-derived embeddings. This finding suggests that (1) the high-quality representations from LLMs effectively encode textual nuances, enabling conventional algorithms to achieve competitive even better detection performance directly in the input space; and (2) the added complexity of deep anomaly detectors may be unnecessary when leveraging such high-quality text representation.

    \item When utilizing the embedding from GloVe and BERT, CVDD exhibits a slight advantage over two-stage AD methods including conventional shallow and deep learning-based methods.

    \item No single method universally outperforms others across all datasets. Across all datasets, the average performance of KNN outperforms all others methods.
\end{itemize}

To further evaluate the performance of two-stage AD methods on different text embeddings, we visualize the heatmaps of AUROC(\%) across all datasets in Figure~\ref{fig-auroc-heatmap}. This visualization clearly demonstrates that LLM-derived embeddings outperform conventional embedding techniques (GloVe and BERT) in most cases, with particularly pronounced improvements in sematic anomaly detection tasks.

\subsection{Comparison between embedding-based and prompt-based text anomaly detection}

\begin{tcolorbox}[
  enhanced,
  title={System template used in Prompt-based text anomaly detection},
  colback=gray!2,
  colframe=black!40,
  boxrule=0.6pt,
  arc=2mm,
  left=2mm,right=2mm,top=1.2mm,bottom=1.2mm,
]
\label{system-prompt}
\begin{lstlisting}[style=prompt]
"You are an expert in unsupervised text anomaly detection.\n"
"You will be given several normal reference samples and multiple query texts.\n"
"For each query text, output an anomaly score between 0 and 1.\n\n"

"Anomaly score definition:\n"
"- 0.0 = completely normal, highly consistent with the reference samples\n"
"- 1.0 = definitely abnormal, clearly inconsistent with the reference samples\n"
"- Higher score means more abnormal\n\n"

"A query should receive a higher anomaly score if it shows any noticeable deviation in:\n"
"- topic or domain\n"
"- intent or behavior\n"
"- wording or style\n"
"- semantic meaning\n"
"- logical consistency\n"
"- distribution compared with the normal samples\n\n"

"Output format:\n"
"- Output exactly one score for each query.\n"
"- Each output line must be in the format: [Query i] <score>\n"
"- Do not output any explanation.\n"
"- Example:\n"
"[Query 1] 0.12\n"
"[Query 2] 0.83\n"
"[Query 3] 0.47\n\n"

"Here are the normal reference samples:\n\n"
\end{lstlisting}
\end{tcolorbox}

In this section, we designed an experiment to compare the detection performance between embedding-based pipelines and prompt-based LLM approaches. Following the setting of our benchmark, we employed Mistral-7B and Llama3-8B as the base models. For prompt-based approaches, the instruction-fine-tuned versions, Mistral-7B-Instruct-v0.2\footnote{https://modelscope.cn/models/AI-ModelScope/Mistral-7B-Instruct-v0.2} and Meta-Llama-3-8B-Instruct\footnote{https://modelscope.cn/models/LLM-Research/Meta-Llama-3-8B-Instruct}, are selected as test objects. To control the experimental scale, we sampled 1,000 samples for training and 2,000 samples for test per dataset and selected KNN and AE as the anomaly detection algorithms in this comparison based on their superior performance in our benchmark. For prompt-based approaches, due to the practical constraints of the maximum context length of LLMs and GPU memory, we had to conduct batch reasoning by randomly sampling 100 normal texts as prompt contents and inferring 10 query texts per batch. The overall system prompt is provided in Table~\ref{system-prompt}.

The corresponding results are presented in Table~\ref{tab-comp-auroc} and Table~\ref{tab-comp-auprc}. As shown, the embedding-based methods show a significant performance advantage on most detection tasks compared to the prompt-based LLM approaches. On the other hand, we also note that prompt-based methods achieve superior performance on IMDB and SST-2. We argue that these datasets correspond to semantic polarity tasks, where anomalies align with high-level linguistic concepts that can be directly inferred by pretrained LLMs. In contrast, embedding-based methods rely on distributional structure and are less sensitive when semantic differences do not correspond to clear geometric separation in the latent space. In summary, prompt-based methods excel at understanding sentiment shifts, while embedding-based methods excel at detecting distributional deviations.

\begin{table}[!ht]
\centering
\caption{Performance comparison (AUROC (\%)) across all datasets.}
\label{tab-comp-auroc}
\resizebox{\textwidth}{!}{
\begin{tabular}{c|c|c|c|c|c|c|c|c|c|c|c|c|c|c}
\toprule
\multicolumn{2}{c|}{\textbf{Method}} &
{\textbf{20news.}} & {\textbf{Reuters}} & {\textbf{IMDB}} & {\textbf{SST2}} &
{\textbf{SMS-Spam}} & {\textbf{Enron}} & {\textbf{WOS}} &
{\textbf{DBp.0}} & {\textbf{DBp.1}} & {\textbf{DBp.2}} &
{\textbf{DBp.3}} & {\textbf{DBp.4}} & {\textbf{Avg}} \\
\midrule
\multirow{2}{*}{Prompt-based } & Mistral-7B-Instruct-v0.2 & 31.25 & 55.96 & \textbf{73.13} & \textbf{82.10} & 72.92 & 57.76 & 43.27 & 44.70 & 56.82 & 46.39 & 50.18 & 56.39 & 55.91 \\
& Llama3-8B-Instruct & 44.14 & 50.00 & 50.00 & 80.86 & 86.48 & 52.38 & 50.00 & 45.78 & 51.73 & 45.92 & 49.41 & 48.18 & 54.57 \\
\midrule
\multirow{4}{*}{Embedding-based} & KNN(Mistral-7B-supervised) & 51.45 & 92.99 & 54.68 & 53.73 & 85.47 & 77.88 & \textbf{58.50} & 85.01 & 94.83 & 97.49 & 99.88 & 98.02 & 79.16 \\
& AE(Mistral-7B-supervised) & 54.80 & \textbf{95.83} & 53.98 & 55.57 & 87.61 & 83.20 & 56.36 & \textbf{89.07} & \textbf{97.84} & \textbf{99.05} & \textbf{99.96} & \textbf{99.67} & \textbf{81.08} \\
& KNN(Llama3-8B-supervised) & 58.93 & 88.30 & 50.95 & 48.46 & \textbf{92.83} & 81.65 & 54.54 & 72.55 & 92.36 & 88.77 & 94.98 & 96.81 & 76.76 \\
& AE(Llama3-8B-supervised) & \textbf{61.91} & 91.55 & 50.41 & 51.50 & 92.73 & \textbf{84.06} & 54.69 & 78.32 & 95.69 & 92.13 & 97.76 & 96.87 & 78.97 \\
\bottomrule
\end{tabular}}
\end{table}

\begin{table}[!ht]
\centering
\caption{Performance comparison (AUPRC (\%)) across all datasets.}
\label{tab-comp-auprc}
\resizebox{\textwidth}{!}{
\begin{tabular}{c|c|c|c|c|c|c|c|c|c|c|c|c|c|c}
\toprule
\multicolumn{2}{c|}{\textbf{Method}} &
{\textbf{20news.}} & {\textbf{Reuters}} & {\textbf{IMDB}} & {\textbf{SST2}} &
{\textbf{SMS-Spam}} & {\textbf{Enron}} & {\textbf{WOS}} &
{\textbf{DBp.0}} & {\textbf{DBp.1}} & {\textbf{DBp.2}} &
{\textbf{DBp.3}} & {\textbf{DBp.4}} & {\textbf{Avg}} \\
\midrule
\multirow{2}{*}{Prompt-based } & Mistral-7B-Instruct-v0.2 & 38.90 & 54.53 & \textbf{69.58} & 78.91 & 63.59 & 59.39 & 44.71 & 49.32 & 58.53 & 49.72 & 52.41 & 53.32 & 56.07 \\
& LLama3-8B-Instruct & 44.51 & 50.00 & 50.00 & \textbf{79.37} & 79.30 & 52.41 & 50.00 & 46.77 & 51.63 & 46.92 & 51.07 & 48.85 & 54.23
 \\
\midrule
\multirow{4}{*}{Embedding-based} & KNN(Mistral-7B-supervised) & 49.37 & 93.63 & 54.18 & 54.12 & 78.75 & 75.11 & \textbf{56.08} & 85.35 & 96.13 & 97.92 & 99.89 & 98.64 & 78.26\\
& AE(Mistral-7B-supervised) & 53.02 & \textbf{96.57} & 54.11 & 54.69 & 81.94 & 80.65 & 53.76 & \textbf{88.34} & \textbf{97.96} & \textbf{99.09} & \textbf{99.96} & \textbf{99.67} & \textbf{79.98} \\
& KNN(Llama3-8B-supervised) & 56.12 & 89.46 & 49.44 & 49.55 & 91.14 & 79.18 & 51.59 & 74.62 & 93.67 & 89.24 & 95.56 & 95.62 & 76.26 \\
& AE(Llama3-8B-supervised) & \textbf{60.56} & 92.33 & 49.61 & 51.48 & \textbf{90.63} & \textbf{81.49} & 53.14 & 78.52 & 96.02 & 92.7 & 97.95 & 95.85 & 78.35 \\
\bottomrule
\end{tabular}}
\end{table}

\section{Low-rank analysis and prediction}

\begin{figure}[h!]
    \centering
    \includegraphics[width=\linewidth]{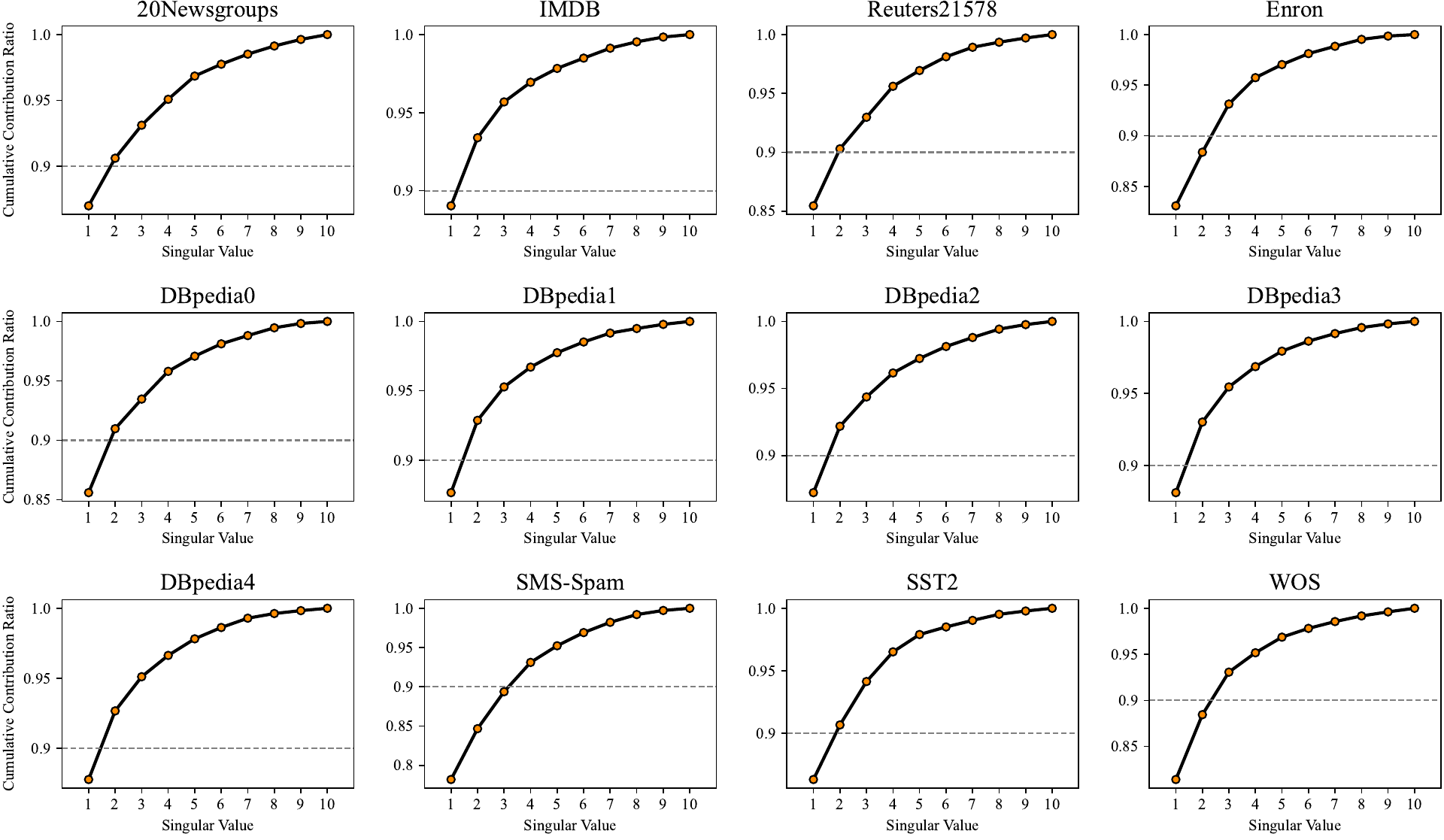}
    \caption{The cumulation contribution ratio of singular value of detection performance (AUROC) across all datasets.}
    \label{fig-singular-ccr}
\end{figure}

To systematically analyze the performance properties across datasets, we conduct singular value decomposition (SVD) on the performance matrices (AUROC on each dataset) (Table~\ref{tab-results-20newsgroups} to Table~\ref{tab-results-dbpedia-4}) where each performance matrix (or table) is a holistic evaluation of detection performance on one dataset across all text embeddings (rows) and AD algorithms (columns). Let $\mathbf{P}\in\mathbb{R}^{m\times n}$ be the performance matrix and $\sigma_i$ denotes the $i$-th singular value, where $\sigma_1 \geq \sigma_2 \geq \cdots \geq \sigma_n > 0$, $m$ denotes the number of embeddings, $n$ denotes the number of AD algorithms, and $n\leq m$. Figure~\ref{fig-singular-ccr} presents the cumulative contribution ratio of the singular values across all datasets, where the cumulative contribution ratio ($ccr$) is defined by $ccr(j) = (\sum_{i=1}^j \sigma_i) / (\sum_{i=1}^n \sigma_i)$.

As evidenced by Figure~\ref{fig-singular-ccr}, the cumulative contribution ratio of the first two singular values ($ccr(2)$) exceeds 0.90 on most datasets. This demonstrates that the AUROC matrices possess strongly low-rank characteristics. This finding has two important implications: (1) the detection performance of novel text datasets or anomaly detection methods can be reliably predicted using only a subset of performance measurements, and (2) this property enables an efficient strategy for rapid model evaluation (embedding evaluation) and selection in practical applications.

\begin{figure}
    \centering
    \includegraphics[width=0.80\linewidth]{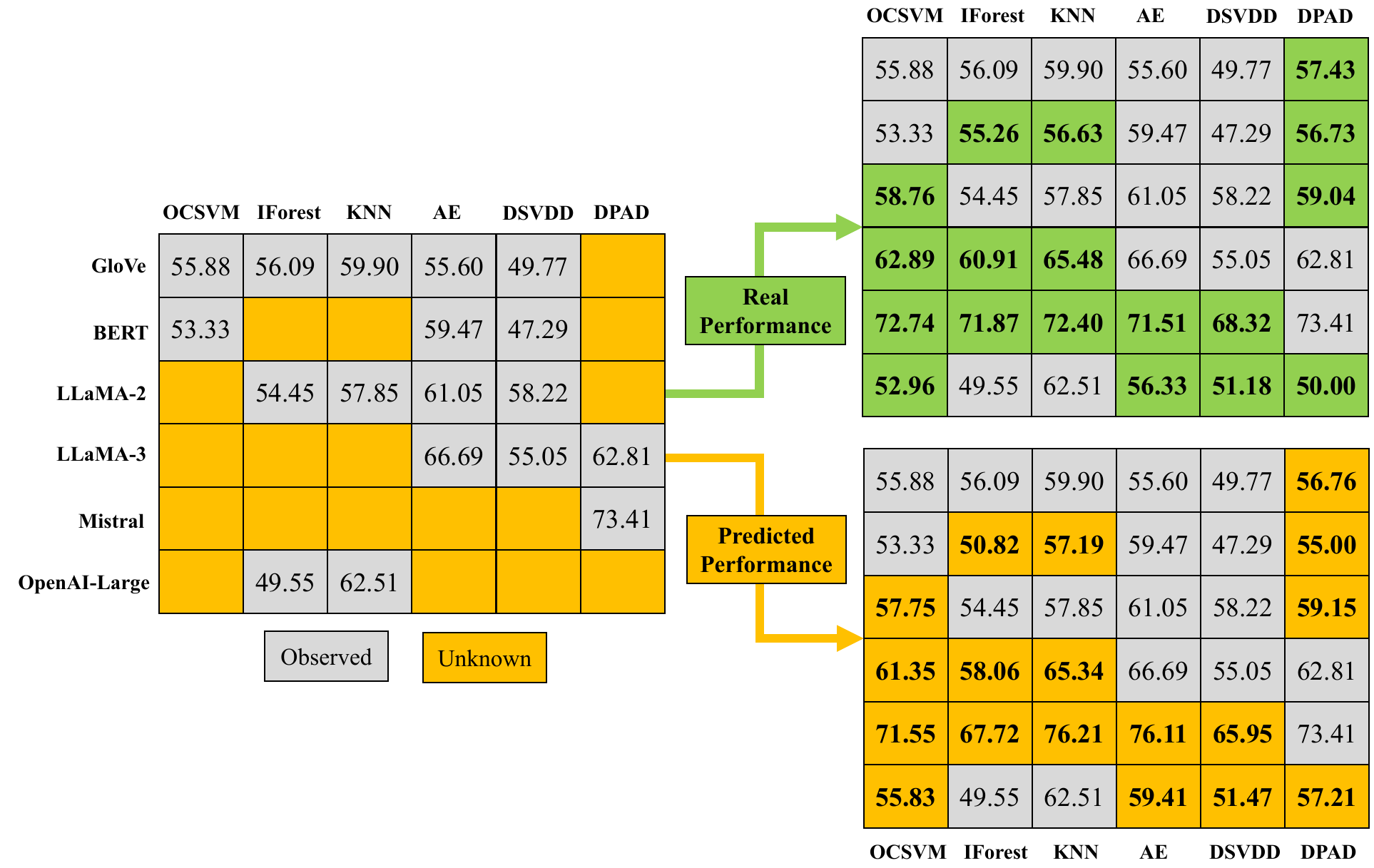}
    \caption{The visualization of AUROC performance prediction on the 20Newsgroups dataset, where the missing rate is 0.5. The prediction process is completed in 1.88 seconds, while the actual performance evaluation requires 5099.50 seconds. }
    \label{fig-prediction}
\end{figure}

To further verify the low-rank characteristics of the AUROC matrices (Table~\ref{tab-results-20newsgroups} to Table~\ref{tab-results-dbpedia-4}), we construct the matrix completion task using the performance matrices. Specifically, for every AUROC matrix, we construct matrix completion \cite{candes2012exact,fan2019factor} tasks by MCAR (missing completely at random) mechanism with missing rate $\in \{0.5, 0.6, 0.7\}$. We use matrix factorization and non-convex optimization techniques~\cite{chi2018low} to recover the missed entries of the performance matrices. Formally, we consider a rank-constrained least-squares problem:
\begin{equation}
    \underset{\bold{\Phi} \in \mathbb{R}^{m \times n}}{\min} ~\Vert \mathcal{P}_{\Omega} (\bold{\Phi} - \mathbf{P}) \Vert_F^2, ~~~~\text{s.t.}~~\text{rank}(\bold{\Phi}) \leq r,
    \label{eq-1}
\end{equation}
where $\mathbf{P} \in \mathbb{R}^{m \times n}$ denotes the performance matrix with $m$ rows (text embeddings) and $n$ columns (AD algorithms), $\Omega$ consists of the locations of observed entries and the observation operator $\mathcal{P}_{\Omega}: \mathbb{R}^{m \times n} \rightarrow \mathbb{R}^{m \times n}$ as
\begin{equation}
    [\mathcal{P}_{\Omega}(\mathbf{P})]_{ij} = 
        \begin{cases} 
            \mathbf{P}_{ij}, & (i,j) \in \Omega \\
            0, & \text{otherwise}
        \end{cases}.
\end{equation}

Invoking the low-rank factorization $\bold{\Phi} = \mathbf{U}\mathbf{V}^T$, where $\mathbf{U} \in \mathbb{R}^{m \times r}$ and $\mathbf{V} \in \mathbb{R}^{n \times r}$, we can rewrite \eqref{eq-1} as an unconstrained optimization problem:
\begin{equation}
    \underset{\mathbf{U}, \mathbf{V}}{\min} ~\Vert \mathcal{P}_{\Omega} (\mathbf{U}\mathbf{V}^T - \mathbf{P}) \Vert_F^2.
    \label{eq-2}
\end{equation}

In our experiments, we set $r=1$ and utilize the singular value decomposition technique to initialize the $\mathbf{U}=\mathbf{U}_0\bold{\Sigma}_0^{1/2}$ and $\mathbf{V}=\mathbf{V}_0\bold{\Sigma}_0^{1/2}$, where $\mathbf{U}_0\bold{\Sigma}_0\mathbf{V}_0^T$ is the best rank-r approximation of $\mathcal{P}_{\Omega}(\mathbf{P})$. The normalized Mean Absolute Percentage Error (MAPE), as defined in Equation~\eqref{eq-mape}, is employed to quantify the accuracy of the prediction.

\begin{equation}
     \text{MAPE} = \frac{1}{k} \sum_{i=1}^{k} \left| \frac{p_i - \hat{p}_i}{p_i} \right|,
     \label{eq-mape}
\end{equation}
where $p_i, \hat{p}_i \in \mathbb{R}$ denote the missing (or unknown) values of detection performance and the corresponding recovered (or predicted) values, respectively. The $k$ denotes the number of unknown values, and the missing rate is $\text{mr} = k/mn$.

The recovery results are reported in Table~\ref{tab-low-rank-completion}, where most of the recovery errors are below 0.1. The results verify the feasibility of a rapid model evaluation (embedding evaluation) and selection in practical applications based on the empirical evaluations in this benchmark. Take the 20Newsgroups dataset as an example, we use 50\% entries of the AUROC performance matrix (see Table~\ref{tab-results-20newsgroups}) to predict other values and show the result of six AD methods on six embeddings in Figure~\ref{fig-prediction}. We can see that the predictions are close to the real performance values. That means, a rapid and relatively accurate prediction of the performance of the new pairs (text embedding, AD method) can be obtained based on some observed results from this benchmark, which guides an efficient and reliable model evaluation (embedding evaluation) and selection.

\begin{table}[h!]
    \centering
        \caption{Recovery performance (MAPE) across all datasets in the setting of MCAR. Note that `mr' denotes missing rate.}
        \label{tab-low-rank-completion}
        \resizebox{\textwidth}{!}{
        \begin{tabular}{lcccccccccccc}
        \toprule
            & 20News. & IMDB & Enton & Reuters. & DBp.0 & DBp.1 & DBp.2 & DBp.3 & DBp.4 & SMS-SPAM & SST2 & WOS \\
        \midrule
        mr=0.5 & 0.0348 & 0.0254 & 0.0419 & 0.0380 & 0.0357 & 0.0375 & 0.0301 & 0.0279 & 0.0286 & 0.0698 & 0.0336 & 0.0570 \\
        mr=0.6 & 0.0431 & 0.0460 & 0.0624 & 0.0404 & 0.0433 & 0.0452 & 0.0438 & 0.0344 & 0.0433 & 0.0828 & 0.0482 & 0.0749 \\
        mr=0.7 & 0.0746 & 0.0535 & 0.1233 & 0.0684 & 0.0787 & 0.0795 & 0.0661 & 0.0589 & 0.0753 & 0.1534 & 0.0771 & 0.1282 \\
        \bottomrule
    \end{tabular}
    }
\end{table}

\section{Conclusion}

In this work, we present Text-ADBench, a comprehensive benchmark for text anomaly detection, which both considers shallow and deep AD approaches, constructs abundant two-stage text AD methods by concatenating LLMs with different pooling strategies and estimates the detection accuracy across eight datasets. Experimental results reveal that \textit{\textbf{(i)}} LLM-based embeddings boost the detection performance for these two-stage AD methods and ``EOS'' pooling exhibits significant advantages over ``Mean'' and ``Weighted Mean'' in most cases, and \textit{\textbf{(ii)}} deep learning-based detectors (AE, DSVDD, DPAD) exhibit no advantage over conventional ``shallow'' algorithms (OCSVM, IForest, KNN, LOF, KDE) when using LLM-derived embeddings, and \textit{\textbf{(iii)}} the average performance of KNN outperforms all other methods, although no single method universally outperforms other across all datasets. Furthermore, we analyze the performance matrices across datasets and find strongly low-rank characteristics, which enable an efficient strategy for rapid model evaluation (embedding evaluation) and selection in practical applications by using the historical estimation from Text-ADBench.

\bibliographystyle{unsrtnat}
\bibliography{references}

\newpage

\begin{table}[h!]
    \centering
    \caption{Average AUROC(\%) on 20Newsgroups. The best results are marked in \textbf{bold}.}
    \label{tab-results-20newsgroups}
    \resizebox{\textwidth}{!}{
\begin{tabular}{c|c|c|c|c|c|c|c|c|c|c|c|c}
    \toprule
    \textbf{Embedding} & \textbf{Pooling} & OCSVM & IForest & LOF & PCA & KNN & KDE & ECOD & AE & DSVDD & DPAD & Avg\\
    \midrule
    GloVe & Mean & 55.88 & 56.09 & 60.16 & 56.36 & 59.90 & 55.99 & 55.42 & 55.60 & 49.77 & 57.43 & 56.26\\
    \midrule
    \multirow{2}{*}{BERT} & CLS & 53.33 & 55.26 & 64.69 & 55.50 & 56.63 & 54.03 & 54.34 & 59.47 & 47.29 & 56.73 & 55.73\\
    & Mean & 52.80 & 55.17 & 58.86 & 54.21 & 55.61 & 53.90 & 52.41 & 54.84 & 54.49 & 56.64 & 54.89\\
    \midrule
    \multirow{3}{*}{\makecell{LLaMA-2\\(mntp)}} & EOS & 50.34 & 48.51 & 44.65 & 47.76 & 49.60 & 47.07 & 47.84 & 46.56 & 45.08 & 46.57 & 47.40\\
    & Mean & 57.73 & 51.73 & 51.06 & 53.31 & 49.92 & 54.96 & 52.76 & 49.34 & 49.99 & 48.66 & 51.95\\
    & Weighted Mean & 57.49 & 51.05 & 50.27 & 52.93 & 49.36 & 54.23 & 52.41 & 47.51  & 48.49 & 47.79 & 51.15\\
    \midrule
    \multirow{3}{*}{\makecell{LLaMA-2\\(mntp-unsup-simcse)}} & EOS & 62.87 & 61.58 & 58.53 & 62.84 & 58.16 & 62.38 & 58.74 & 55.69 & 55.60 & 56.63 & 59.30\\
\bottomrule
    \end{tabular}
    }
\end{table}

\begin{table}[h!]
    \centering
    \caption{Average AUROC(\%) on Enron.}
    \label{tab-results-enron}
    \resizebox{\textwidth}{!}{
    \begin{tabular}{c|c|c|c|c|c|c|c|c|c|c|c|c}
    \toprule
    \textbf{Embedding} & \textbf{Pooling} & OCSVM & IForest & LOF & PCA & KNN & KDE & ECOD & AE & DSVDD & DPAD & Avg\\
    \midrule
    GloVe & Mean & 63.12 & 62.45 & 67.86 & 62.64 & 73.63 & 63.57 & 58.58 & 69.41 & 56.91 & 73.99 & 65.22\\
    \midrule
    \multirow{2}{*}{BERT} & CLS & 53.74 & 53.72 & 76.81 & 54.28 & 68.02 & 53.49 & 52.78 & 63.68 & 44.82 & 79.74 & 60.11 \\
    & Mean & 53.26 & 54.83 & 76.18 & 55.01 & 61.86 & 54.88 & 51.93 & 59.28 & 54.60 & 76.96 &  59.88\\
    \midrule
    \multirow{3}{*}{\makecell{LLaMA-2\\(mntp)}} & EOS & 49.45 & 51.32 & 63.05 & 50.81 & 62.33 & 51.07 & 49.14 & 62.76 & 50.53 & 86.86 &  57.73\\
    & Mean & 53.25 & 57.01 & 75.37 & 54.82 & 65.48 & 53.28 & 52.89 & 63.47 & 52.50 & 75.49 &  60.36\\
    & Weighted Mean & 52.72 & 55.04 & 73.45 & 53.73 & 62.56 & 52.70 & 52.25 & 61.04 & 53.65 & 74.64 &  59.18\\
    \midrule
    \multirow{3}{*}{\makecell{LLaMA-2\\(mntp-unsup-simcse)}} & EOS & 79.37 & 70.29 & 93.97 & 73.96 & 90.00 & 73.96 & 69.56 & \textbf{98.72} & 82.24 & 96.87 &  82.89\\
\bottomrule
    \end{tabular}
    }
\end{table}

\begin{table}[h!]
    \centering
    \caption{Average AUROC(\%) on IMDB.}
    \label{tab-results-imdb}
    \resizebox{\textwidth}{!}{
    \begin{tabular}{c|c|c|c|c|c|c|c|c|c|c|c|c}
    \toprule
    \textbf{Embedding} & \textbf{Pooling} & OCSVM & IForest & LOF & PCA & KNN & KDE & ECOD & AE & DSVDD & DPAD & Avg\\
    \midrule
    GloVe & Mean & 46.35 & 46.76 & 49.31 & 46.51 & 46.74 & 46.37 & 42.95 & 48.66 & 47.72 & 48.31 & 46.97 \\
    \midrule
    \multirow{2}{*}{BERT} & CLS & 44.33 & 45.77 & 45.66 & 44.46 & 46.31 & 44.40 & 40.95 & 45.12 & 48.91 & 46.76 & 45.27 \\
    & Mean & 45.11 & 45.34 & 47.75 & 44.59 & 47.65 & 44.29 & 39.24 & 46.87 & 46.19 & 46.96 & 45.40 \\
    \midrule
    \multirow{3}{*}{\makecell{LLaMA-2\\(mntp)}} & EOS & 48.36 & 48.33 & 50.26 & 48.33 & 49.32 & 48.33 & \textbf{48.10} & 49.20 & 48.90 & 49.83 & 48.90 \\
    & Mean & 46.58 & 47.22 & 49.44 & 46.70 & 49.39 & 47.03 & 46.07 & 49.45 & 48.34 & 50.11 & 48.03 \\
    & Weighted Mean & 46.78 & 47.37 & 50.07 & 46.81 & 49.53 & 47.15 & 46.22 & 49.41 & 48.55 & 50.16 & 48.21 \\
    \midrule
    \multirow{3}{*}{\makecell{LLaMA-2\\(mntp-unsup-simcse)}} & EOS & 46.81 & 45.85 & 54.24 & 45.19 & 54.25 & 45.31 & 36.69 & 55.03 & 47.54 & 50.96 & 48.19 \\
\bottomrule
    \end{tabular}
    }
\end{table}

\begin{table}[h!]
    \centering
    \caption{Average AUROC(\%) on Reuters21578.}
    \label{tab-results-reuters}
    \resizebox{\textwidth}{!}{
    \begin{tabular}{c|c|c|c|c|c|c|c|c|c|c|c|c}
    \toprule
    \textbf{Embedding} & \textbf{Pooling} & OCSVM & IForest & LOF & PCA & KNN & KDE & ECOD & AE & DSVDD & DPAD & Avg\\
    \midrule
    GloVe & Mean & 76.85 & 78.21 & 84.54 & 77.77 & 82.75 & 75.96 & 64.57 & 80.93 & 57.92 & 83.08 & 76.26 \\
    \midrule
    \multirow{2}{*}{BERT} & CLS & 65.90 & 58.08 & 76.69 & 58.43 & 52.54 & 60.90 & 49.07 & 63.65 & 54.05 & 61.18  &  60.05\\
    & Mean & 80.74 & 80.79 & 89.45 & 80.45 & 83.61 & 78.06 & 66.28 & 83.64 & 71.45 & 88.87 &  80.33\\
    \midrule
    \multirow{3}{*}{\makecell{LLaMA-2\\(mntp)}} & EOS & 62.17 & 64.19 & 59.43 & 65.07 & 67.87 & 63.69 & 59.71 & 69.37 & 63.61 & 73.05 &  64.82\\
    & Mean & 56.63 & 64.05 & 61.33 & 60.15 & 65.36 & 55.25 & 51.87 & 72.80 & 57.44 & 66.75 & 61.16 \\
    & Weighted Mean & 56.80 & 64.30 & 61.29 & 60.16 & 65.15 & 56.23 & 53.06 & 67.89 & 58.62 & 67.48 & 61.10 \\
    \midrule
    \multirow{3}{*}{\makecell{LLaMA-2\\(mntp-unsup-simcse)}} & EOS & 95.38 & 91.17 & 86.94 & 95.37 & 95.48 & 95.03 & 87.49 & 96.42 & 87.06 & \textbf{94.56} &  \textbf{92.49}\\
\bottomrule
    \end{tabular}
    }
\end{table}

\begin{table}[h!]
    \centering
    \caption{Average AUROC(\%) on SMS-SPAM.}
    \label{tab-results-sms_spam}
    \resizebox{\textwidth}{!}{
    \begin{tabular}{c|c|c|c|c|c|c|c|c|c|c|c|c}
    \toprule
    \textbf{Embedding} & \textbf{Pooling} & OCSVM & IForest & LOF & PCA & KNN & KDE & ECOD & AE & DSVDD & DPAD & Avg\\
    \midrule
    GloVe & Mean & 48.02 & 33.43 & 44.65 & 34.60 & 35.43 & 41.38 & 32.81 & 31.25 & 64.28 & 36.78 & 40.26 \\
    \midrule
    \multirow{2}{*}{BERT} & CLS & 61.94 & 33.87 & 43.46 & 37.87 & 24.09 & 53.32 & 30.65 & 32.28 & 64.43 & 36.44 & 41.84 \\
    & Mean & 54.87 & 46.89 & 56.05 & 50.44 & 33.36 & 51.41 & 32.27 & 42.61 & 89.43 & 60.38 & 51.77 \\
    \midrule
    \multirow{3}{*}{\makecell{LLaMA-2\\(mntp)}} & EOS & 77.78 & 72.10 & 69.49 & 74.03 & 37.45 & 73.46 & 53.77 & 53.72 & 46.29 & 63.70 & 62.18 \\
    & Mean & 83.29 & 78.76 & 57.48 & 78.45 & 24.03 & 79.76 & 55.17 & 23.46 & 49.16 & 31.14 & 56.07 \\
    & Weighted Mean & 86.29 & 82.84 & 73.59 & 82.85 & 24.68 & 84.22 & 59.15 & 30.42 & 42.61 & 38.52 & 60.52 \\
    \midrule
    \multirow{3}{*}{\makecell{LLaMA-2\\(mntp-unsup-simcse)}} & EOS & 84.47 & 73.75 & 85.16 & 84.62 & 82.78 & 83.84 & 63.55 & 85.56 & 70.46 & 83.84 & 79.80 \\
\bottomrule
    \end{tabular}
    }
\end{table}

\begin{table}[h!]
    \centering
    \caption{Average AUROC(\%) on SST2.}
    \label{tab-results-sst2}
    \resizebox{\textwidth}{!}{
    \begin{tabular}{c|c|c|c|c|c|c|c|c|c|c|c|c}
    \toprule
    \textbf{Embedding} & \textbf{Pooling} & OCSVM & IForest & LOF & PCA & KNN & KDE & ECOD & AE & DSVDD & DPAD & Avg\\
    \midrule
    GloVe & Mean & 46.38 & 44.69 & 58.73 & 44.79 & 48.68 & 45.23 & 43.83 & 49.60 & 53.47 & 54.09 & 48.95 \\
    \midrule
    \multirow{2}{*}{BERT} & CLS & 52.93 & 51.93 & 58.17 & 51.82 & 55.93 & 52.40 & 49.58 & 55.04 & 51.98 & 60.13 & 53.99 \\
    & Mean & 51.94 & 50.73 & 58.66 & 51.45 & 59.00 & 51.32 & 48.39 & 57.39 & 51.35 & 62.26 & 54.25 \\
    \midrule
    \multirow{3}{*}{\makecell{LLaMA-2\\(mntp)}} & EOS & 49.80 & 50.25 & 53.66 & 49.80 & 57.31 & 49.37 & 48.97 & 58.31 & 50.41 & 64.76 & 53.26 \\
    & Mean & 49.30 & 48.99 & 54.87 & 48.44 & 55.45 & 48.39 & 47.83 & 56.74 & 50.29 & 62.60 & 52.29 \\
    & Weighted Mean & 48.75 & 48.94 & 54.63 & 48.18 & 55.68 & 48.04 & 47.49 & 56.41 & 49.73 & 63.92 & 52.18 \\
    \midrule
    \multirow{3}{*}{\makecell{LLaMA-2\\(mntp-unsup-simcse)}} & EOS & 59.54 & 55.85 & 69.28 & 60.01 & 75.26 & 60.30 & 52.98 & 79.73 & 59.35 & 73.34 & 64.56 \\
\bottomrule
    \end{tabular}
    }
\end{table}

\begin{table}[h!]
    \centering
    \caption{Average AUROC(\%) on WOS.}
    \label{tab-results-wos}
    \resizebox{\textwidth}{!}{
    \begin{tabular}{c|c|c|c|c|c|c|c|c|c|c|c|c}
    \toprule
    \textbf{Embedding} & \textbf{Pooling} & OCSVM & IForest & LOF & PCA & KNN & KDE & ECOD & AE & DSVDD & DPAD & Avg\\
    \midrule
    GloVe & Mean & 52.49 & 38.61 & 64.93 & 39.30 & 33.74 & 47.93 & 36.11 & 40.26 & 56.32 & 49.69 & 45.94 \\
    \midrule
    \multirow{2}{*}{BERT} & CLS & 47.15 & 47.21 & 58.02 & 45.24 & 46.23 & 45.26 & 42.84 & 46.73 & 47.94 & 49.57 & 47.62\\
    & Mean & 53.09 & 50.35 & 59.02 & 49.05 & 37.77 & 49.52 & 43.80 & 41.25 & 49.32 & 48.70 &  48.19\\
    \midrule
    \multirow{3}{*}{\makecell{LLaMA-2\\(mntp)}} & EOS & 47.46 & 46.17 & 49.46 & 46.92 & 50.92 & 45.99 & 46.68 & 51.51 & 48.03 & 51.74 &  48.49\\
    & Mean & 51.40 & 48.20 & 49.87 & 47.89 & 44.30 & 49.80 & 45.72 & 41.10 & 46.66 & 46.57 &  47.15\\
    & Weighted Mean & 50.81 & 46.20 & 49.01 & 47.56 & 43.84 & 49.00 & 45.80 & 40.44 & 47.12 & 46.34 & 46.61 \\
    \midrule
    \multirow{3}{*}{\makecell{LLaMA-2\\(mntp-unsup-simcse)}} & EOS & 59.93 & 55.46 & 60.92 & 59.98 & 63.82 & 60.33 & 52.78 & 62.55 & 56.07 & 60.37 &  59.22\\
\bottomrule
    \end{tabular}
    }
\end{table}

\begin{table}[h!]
    \centering
    \caption{Average AUROC(\%) on DBpedia-0.}
    \label{tab-results-dbpedia-0}
    \resizebox{\textwidth}{!}{
    \begin{tabular}{c|c|c|c|c|c|c|c|c|c|c|c|c}
    \toprule
    \textbf{Embedding} & \textbf{Pooling} & OCSVM & IForest & LOF & PCA & KNN & KDE & ECOD & AE & DSVDD & DPAD & Avg\\
    \midrule
    GloVe & Mean & 82.33 & 77.03 & 86.25 & 78.19 & 79.83 & 81.04 & 65.91 & 79.30 & 56.61 & 82.42 & 76.89\\
    \midrule
    \multirow{2}{*}{BERT} & CLS & 84.94 & 83.48 & 94.62 & 84.91 & 88.83 & 83.56 & 78.23 & 91.72 & 73.72 & 90.05 &  85.41\\
    & Mean & 81.63 & 78.96 & 91.35 & 80.39 & 87.41 & 80.11 & 71.79 & 90.65 & 81.18 & 90.67 &  83.41\\
    \midrule
    \multirow{3}{*}{\makecell{LLaMA-2\\(mntp)}} & EOS & 66.76 & 65.36 & 73.55 & 66.98 & 66.16 & 65.87 & 62.14 & 69.01 & 59.94 & 66.76 &  66.25\\
    & Mean & 71.12 & 67.53 & 75.38 & 68.81 & 64.65 & 68.76 & 60.57 & 64.77 & 64.68 & 63.27 &  66.95\\
    & Weighted Mean & 68.45 & 65.79 & 76.16 & 66.89 & 64.00 & 66.50 & 60.16 & 61.86 & 61.99 & 62.72 &  65.45\\
    \midrule
    \multirow{3}{*}{\makecell{LLaMA-2\\(mntp-unsup-simcse)}} & EOS & 95.81 & \textbf{84.70} & 96.94 & 95.58 & 97.19 & 95.72 & 86.18 & 96.90 & 84.03 & \textbf{93.53} &  \textbf{92.66}\\
\bottomrule
    \end{tabular}
    }
\end{table}

\begin{table}[h!]
    \centering
    \caption{Average AUROC(\%) on DBpedia-1.}
    \label{tab-results-dbpedia-1}
    \resizebox{\textwidth}{!}{
    \begin{tabular}{c|c|c|c|c|c|c|c|c|c|c|c|c}
    \toprule
    \textbf{Embedding} & \textbf{Pooling} & OCSVM & IForest & LOF & PCA & KNN & KDE & ECOD & AE & DSVDD & DPAD & Avg\\
    \midrule
    GloVe & Mean & 91.59 & 88.37 & 96.57 & 90.11 & 89.22 & 90.89 & 76.42 & 89.48 & 68.34 & 96.02 & 87.70 \\
    \midrule
    \multirow{2}{*}{BERT} & CLS & 85.12 & 82.64 & 96.96 & 84.38 & 90.72 & 82.31 & 76.57 & 93.64 & 75.22 & 92.02 & 85.69 \\
    & Mean & 82.06 & 77.26 & 95.04 & 80.88 & 92.76 & 79.44 & 70.92 & 94.50 & 91.28 & 95.08 & 85.92 \\
    \midrule
    \multirow{3}{*}{\makecell{LLaMA-2\\(mntp)}} & EOS & 72.70 & 70.78 & 77.46 & 72.85 & 73.51 & 71.59 & 67.31 & 77.05 & 64.28 & 75.58 & 72.31 \\
    & Mean & 78.29 & 77.44 & 84.46 & 77.84 & 73.07 & 75.90 & 67.14 & 71.41 & 69.54 & 70.32 & 74.54 \\
    & Weighted Mean & 76.40 & 74.59 & 84.16 & 75.83 & 72.22 & 74.01 & 67.42 & 71.15 & 66.28 & 70.59 & 73.26 \\
    \midrule
    \multirow{3}{*}{\makecell{LLaMA-2\\(mntp-unsup-simcse)}} & EOS & 99.36 & 94.23 & 98.87 & 99.29 & 99.57 & 99.17 & 95.50 & 99.68 & 95.37 & 98.98 &  \textbf{98.00}\\
\bottomrule
    \end{tabular}
    }
\end{table}

\begin{table}[h!]
    \centering
    \caption{Average AUROC(\%) on DBpedia-2.}
    \label{tab-results-dbpedia-2}
    \resizebox{\textwidth}{!}{
    \begin{tabular}{c|c|c|c|c|c|c|c|c|c|c|c|c}
    \toprule
    \textbf{Embedding} & \textbf{Pooling} & OCSVM & IForest & LOF & PCA & KNN & KDE & ECOD & AE & DSVDD & DPAD & Avg\\
    \midrule
    GloVe & Mean & 85.27 & 83.18 & 87.29 & 83.39 & 83.86 & 84.03 & 74.83 & 83.76 & 57.31 & 87.38 & 81.03 \\
    \midrule
    \multirow{2}{*}{BERT} & CLS & 74.38 & 72.29 & 93.54 & 73.46 & 81.22 & 71.48 & 67.47 & 86.61 & 69.06 & 83.49 & 77.30 \\
    & Mean & 80.26 & 76.38 & 92.97 & 78.62 & 89.61 & 77.28 & 71.73 & 93.38 & 84.91 & 91.41 & 83.66 \\
    \midrule
    \multirow{3}{*}{\makecell{LLaMA-2\\(mntp)}} & EOS & 66.84 & 63.97 & 73.12 & 66.84 & 65.10 & 65.68 & 60.01 & 66.17 & 61.51 & 61.69 & 65.09 \\
    & Mean & 77.94 & 76.14 & 84.42 & 76.70 & 72.00 & 74.73 & 67.44 & 65.56 & 67.48 & 66.92 & 72.93 \\
    & Weighted Mean & 75.68 & 74.90 & 83.16 & 74.45 & 70.01 & 72.54 & 65.22 & 64.08 & 65.20 & 64.85 & 71.01 \\
    \midrule
    \multirow{3}{*}{\makecell{LLaMA-2\\(mntp-unsup-simcse)}} & EOS & 91.55 & 76.95 & 95.02 & 89.97 & 93.93 & 89.72 & 80.26 & 96.38 & 82.01 & 92.33 & 88.81 \\
\bottomrule
    \end{tabular}
    }
\end{table}

\begin{table}[h!]
    \centering
    \caption{Average AUROC(\%) on DBpedia-3.}
    \label{tab-results-dbpedia-3}
    \resizebox{\textwidth}{!}{
    \begin{tabular}{c|c|c|c|c|c|c|c|c|c|c|c|c}
    \toprule
    \textbf{Embedding} & \textbf{Pooling} & OCSVM & IForest & LOF & PCA & KNN & KDE & ECOD & AE & DSVDD & DPAD & Avg\\
    \midrule
    GloVe & Mean & 87.97 & 82.11 & 97.15 & 83.69 & 92.02 & 85.17 & 69.89 & 91.06 & 66.49 & 95.19 & 85.07 \\
    \midrule
    \multirow{2}{*}{BERT} & CLS & 67.87 & 63.44 & 94.62 & 64.36 & 85.11 & 63.13 & 51.13 & 87.72 & 79.72 & 86.22 &  74.33\\
    & Mean & 75.88 & 72.88 & 96.42 & 70.94 & 95.96 & 69.22 & 56.43 & 96.35 & 93.83 & 96.92 & 82.48 \\
    \midrule
    \multirow{3}{*}{\makecell{LLaMA-2\\(mntp)}} & EOS & 69.63 & 68.89 & 80.81 & 70.94 & 79.30 & 69.28 & 61.81 & 77.33 & 68.06 & 78.44 & 72.45 \\
    & Mean & 81.68 & 82.60 & 90.70 & 82.64 & 82.90 & 80.83 & 71.28 & 78.73 & 75.17 & 77.98 & 80.45 \\
    & Weighted Mean & 78.74 & 81.06 & 88.54 & 79.83 & 81.85 & 78.22 & 68.30 & 77.42 & 73.07 & 77.54 & 78.46 \\
    \midrule
    \multirow{3}{*}{\makecell{LLaMA-2\\(mntp-unsup-simcse)}} & EOS & 98.52 & 91.07 & 99.52 & 98.02 & 99.28 & 97.76 & 93.74 & \textbf{99.71} & 93.79 & 98.69 & \textbf{97.01} \\
\bottomrule
    \end{tabular}
    }
\end{table}

\begin{table}[h!]
    \centering
    \caption{Average AUROC(\%) on DBpedia-4.}
    \label{tab-results-dbpedia-4}
    \resizebox{\textwidth}{!}{
    \begin{tabular}{c|c|c|c|c|c|c|c|c|c|c|c|c}
    \toprule
    \textbf{Embedding} & \textbf{Pooling} & OCSVM & IForest & LOF & PCA & KNN & KDE & ECOD & AE & DSVDD & DPAD & Avg\\
    \midrule
    GloVe & Mean & 86.64 & 82.12 & 92.22 & 83.09 & 85.92 & 84.82 & 70.75 & 86.25 & 61.66 & 91.17 & 82.46 \\
    \midrule
    \multirow{2}{*}{BERT} & CLS & 68.96 & 63.59 & 93.78 & 67.67 & 82.68 & 64.06 & 57.37 & 86.37 & 74.36 & 83.02 & 74.19 \\
    & Mean & 76.02 & 72.27 & 93.65 & 72.75 & 93.23 & 71.38 & 63.34 & 94.07 & 88.91 & 94.38 & 82.00 \\
    \midrule
    \multirow{3}{*}{\makecell{LLaMA-2\\(mntp)}} & EOS & 71.83 & 69.48 & 72.80 & 71.09 & 72.77 & 69.98 & 60.90 & 70.17 & 66.53 & 71.23 & 69.68 \\
    & Mean & 85.18 & 82.91 & 89.40 & 84.03 & 79.83 & 82.68 & 74.62 & 75.26 & 72.24 & 75.87 & 80.20 \\
    & Weighted Mean & 81.99 & 80.23 & 87.43 & 80.91 & 78.56 & 79.13 & 71.28 & 77.33 & 69.73 & 75.90 & 78.25 \\
    \midrule
    \multirow{3}{*}{\makecell{LLaMA-2\\(mntp-unsup-simcse)}} & EOS & 97.43 & 88.44 & 97.72 & 96.30 & 98.78 & 96.16 & 88.80 & 98.91 & 87.72 & 96.91 & 94.72 \\
\bottomrule
    \end{tabular}
    }
\end{table}

\begin{table}[h!]
    \centering
    \caption{Average AUPRC(\%) on DBpedia-0.}
    \label{tab-AUPR-dbpedia-0}
    \resizebox{\textwidth}{!}{
    \begin{tabular}{c|c|c|c|c|c|c|c|c|c|c|c|c}
    \toprule
    \textbf{Embedding} & \textbf{Pooling} & OCSVM & IForest & LOF & PCA & KNN & KDE & ECOD & AE & DSVDD & DPAD & Avg\\
    \midrule
    GloVe & Mean &  75.71 & 70.37 & 81.18 & 71.46 & 78.23 & 74.33 & 57.78 & 76.00 & 49.73 & 79.74 & 71.45 \\
    \midrule
    \multirow{2}{*}{BERT} & CLS &  80.55 & 80.16 & 94.03 & 81.80 & 88.41 & 78.31 & 74.76 & 91.09 & 70.10 & 88.12 & 82.73 \\
    & Mean &  75.26 & 73.05 & 89.84 & 73.91 & 84.82 & 73.11 & 64.81 & 88.09 & 78.47 & 88.49 & 78.99  \\
    \midrule
    \multirow{3}{*}{\makecell{LLaMA-2\\(mntp)}} & EOS & 56.67 & 55.17 & 67.82 & 56.49 & 58.53 & 55.56 & 51.91 & 58.88 & 51.03 & 57.48 & 56.95 \\
    & Mean & 58.97 & 55.13 & 72.02 & 56.11 & 54.86 & 56.08 & 49.52 & 54.88 & 56.32 & 53.51 & 56.74\\
    & Weighted Mean & 57.14 & 54.47 & 73.42 & 54.96 & 54.65 & 54.66 & 49.58 & 53.40 & 54.49 & 53.39 & 56.02\\
    \midrule
    \multirow{3}{*}{\makecell{LLaMA-2\\(mntp-unsup-simcse)}} & EOS &  94.68 & 81.07 & 96.03 & 94.25 & 96.33 & 94.43 & 81.18 & 96.07 & 81.45 & 92.40 & 90.79\\
\bottomrule
    \end{tabular}
    }
\end{table}

\begin{table}[h!]
    \centering
    \caption{Average AUPRC(\%) on DBpedia-1.}
    \label{tab-AUPR-dbpedia-1}
    \resizebox{\textwidth}{!}{
    \begin{tabular}{c|c|c|c|c|c|c|c|c|c|c|c|c}
    \toprule
    \textbf{Embedding} & \textbf{Pooling} & OCSVM & IForest & LOF & PCA & KNN & KDE & ECOD & AE & DSVDD & DPAD & Avg\\
    \midrule
    GloVe & Mean & 86.75 & 82.24 & 94.89 & 84.35 & 88.16 & 85.55 & 66.30 & 87.44 & 59.01 & 94.85 & 82.95\\
    \midrule
    \multirow{2}{*}{BERT} & CLS &  76.32 & 76.46 & 96.13 & 77.89 & 89.81 & 72.45 & 68.87 & 92.51 & 70.44 & 89.13 & 81.00\\
    & Mean &  72.73 & 68.32 & 94.03 & 72.09 & 89.50 & 69.72 & 61.34 & 91.53 & 90.35 & 93.37 & 80.30\\
    \midrule
    \multirow{3}{*}{\makecell{LLaMA-2\\(mntp)}} & EOS &  64.35 & 62.71 & 70.95 & 64.42 & 68.97 & 63.06 & 59.05 & 70.60 & 56.56 & 68.32 & 64.90\\
    & Mean &  68.26 & 68.19 & 82.69 & 68.43 & 65.65 & 64.93 & 58.36 & 65.44 & 62.92 & 62.75 & 66.76\\
    & Weighted Mean &  67.72 & 66.05 & 82.57 & 67.03 & 65.76 & 64.30 & 59.28 & 64.95 & 59.89 & 63.56 & 66.11\\
    \midrule
    \multirow{3}{*}{\makecell{LLaMA-2\\(mntp-unsup-simcse)}} & EOS &  99.19 & 93.01 & 97.80 & 99.10 & 99.42 & 98.93 &  94.22 & 99.53 & 94.74 & 98.74 & 97.47\\
\bottomrule
    \end{tabular}
    }
\end{table}

\begin{table}[h!]
    \centering
    \caption{Average AUPRC(\%) on DBpedia-2.}
    \label{tab-AUPR-dbpedia-2}
    \resizebox{\textwidth}{!}{
    \begin{tabular}{c|c|c|c|c|c|c|c|c|c|c|c|c}
    \toprule
    \textbf{Embedding} & \textbf{Pooling} & OCSVM & IForest & LOF & PCA & KNN & KDE & ECOD & AE & DSVDD & DPAD & Avg\\
    \midrule
    GloVe & Mean & 80.80 & 77.57 & 80.46 & 77.45 & 81.23 & 78.55 & 66.27 & 81.66 & 49.44 & 84.95 & 75.84\\
    \midrule
    \multirow{2}{*}{BERT} & CLS & 65.89 & 65.82 & 92.59 & 66.89 & 80.41 & 62.54 & 60.92 & 85.11 & 64.15 & 79.17 & 72.35\\
    & Mean &  72.93 & 69.45 & 92.30 & 71.37 & 88.32 & 68.77 & 64.33 & 91.91 & 83.29 & 89.92 & 79.26\\
    \midrule
    \multirow{3}{*}{\makecell{LLaMA-2\\(mntp)}} & EOS &  59.23 & 56.31 & 70.00 & 59.26 & 61.23 & 57.93 & 52.93 & 59.69 & 55.93 & 56.55 & 58.91\\
    & Mean &  68.74 & 67.64 & 83.10 & 67.97 & 65.10 & 64.80 & 58.11 & 57.88 & 59.78 & 60.39 & 65.35\\
    & Weighted Mean &  66.79 & 66.50 & 81.56 & 65.82 & 63.97 & 63.06 & 56.63 & 57.28 & 58.34 & 59.57 & 63.95\\
    \midrule
    \multirow{3}{*}{\makecell{LLaMA-2\\(mntp-unsup-simcse)}} & EOS &  90.13 & 72.78 & 94.16 & 88.24 & 93.55 & 87.99 & 75.20 & 95.82 & 79.57 & 91.20 & 86.86\\
\bottomrule
    \end{tabular}
    }
\end{table}

\begin{table}[h!]
    \centering
    \caption{Average AUPRC(\%) on DBpedia-3.}
    \label{tab-AUPR-dbpedia-3}
    \resizebox{\textwidth}{!}{
    \begin{tabular}{c|c|c|c|c|c|c|c|c|c|c|c|c}
    \toprule
    \textbf{Embedding} & \textbf{Pooling} & OCSVM & IForest & LOF & PCA & KNN & KDE & ECOD & AE & DSVDD & DPAD & Avg\\
    \midrule
    GloVe & Mean &  84.38 & 76.84 & 95.78 & 79.43 & 91.92 & 81.01 & 62.90 & 90.64 & 59.40 & 94.51 & 81.68\\
    \midrule
    \multirow{2}{*}{BERT} & CLS &  56.39 & 53.00 & 93.20 & 53.46 & 82.94 & 52.16 & 43.97 & 85.30 & 77.76 & 82.14 & 68.03\\
    & Mean &  66.65 & 64.37 & 95.57 & 62.22 & 94.78 & 59.65 & 49.25 & 95.26 & 93.68 & 96.31 & 77.77\\
    \midrule
    \multirow{3}{*}{\makecell{LLaMA-2\\(mntp)}} & EOS &  63.53 & 62.92 & 75.98 & 64.98 & 76.56 & 62.93 & 56.63 & 71.37 & 64.98 & 74.34 & 67.42\\
    & Mean &  77.93 & 78.58 & 89.65 & 78.81 & 79.86 & 76.45 & 67.97 & 76.15 & 71.52 & 73.70 & 77.06\\
    & Weighted Mean &  75.44 & 76.98 & 86.53 & 76.14 & 78.69 & 74.44 & 65.55 & 74.61 & 69.81 & 73.57 & 75.18\\
    \midrule
    \multirow{3}{*}{\makecell{LLaMA-2\\(mntp-unsup-simcse)}} & EOS &  98.11 & 89.27 & 99.29 & 97.48 & 99.15 & 97.17 & 91.99 & 99.62 & 92.87 & 98.37 & 96.33\\
\bottomrule
    \end{tabular}
    }
\end{table}

\begin{table}[h!]
    \centering
    \caption{Average AUPRC(\%) on DBpedia-4.}
    \label{tab-AUPR-dbpedia-4}
    \resizebox{\textwidth}{!}{
    \begin{tabular}{c|c|c|c|c|c|c|c|c|c|c|c|c}
    \toprule
    \textbf{Embedding} & \textbf{Pooling} & OCSVM & IForest & LOF & PCA & KNN & KDE & ECOD & AE & DSVDD & DPAD & Avg\\
    \midrule
    GloVe & Mean &  81.18 & 76.63 & 88.21 & 77.48 & 84.59 & 78.97 & 63.42 & 85.14 & 53.68 & 89.41 & 77.87\\
    \midrule
    \multirow{2}{*}{BERT} & CLS & 54.95 & 53.48 & 91.63 & 55.20 & 80.75 & 50.06 & 46.91 & 83.62 & 69.06 &  77.50 & 66.32\\
    & Mean &  63.21 & 60.98 & 91.18 & 60.88 & 90.85 & 58.34 & 52.17 & 91.38 & 87.34 & 92.40 & 74.87\\
    \midrule
    \multirow{3}{*}{\makecell{LLaMA-2\\(mntp)}} & EOS &  64.64 & 61.95 & 68.04 & 63.36 & 66.47 & 62.36 & 53.91 & 64.18 & 62.12 & 64.99 & 63.20\\
    & Mean &  77.58 & 75.28 & 87.56 & 76.42 & 74.47 & 73.87 & 64.98 & 70.43 & 65.76 & 70.10 & 73.65\\
    & Weighted Mean &  73.56 & 71.47 & 85.04 & 72.50 & 72.78 & 69.67 & 61.31 & 72.37 & 62.59 & 70.23 & 71.15\\
    \midrule
    \multirow{3}{*}{\makecell{LLaMA-2\\(mntp-unsup-simcse)}} & EOS &  96.38 & 84.68 & 96.88 & 94.76 & 98.37 & 94.60 & 84.25 & 98.50 & 84.37 & 96.00 & 92.88\\
\bottomrule
    \end{tabular}
    }
\end{table}

\begin{table}[h!]
    \centering
    \caption{Average AUPRC(\%) on WOS.}
    \label{tab-AUPR-wos}
    \resizebox{\textwidth}{!}{
    \begin{tabular}{c|c|c|c|c|c|c|c|c|c|c|c|c}
    \toprule
    \textbf{Embedding} & \textbf{Pooling} & OCSVM & IForest & LOF & PCA & KNN & KDE & ECOD & AE & DSVDD & DPAD & Avg\\
    \midrule
    GloVe & Mean &  31.51 & 25.02 & 45.53 & 25.09 & 23.64 & 28.64 & 24.04 & 26.10 & 36.96 & 29.00 & 29.55\\
    \midrule
    \multirow{2}{*}{BERT} & CLS &  29.69 & 29 99 & 38.01 & 29.06 & 30.06 & 28.99 & 27.90 & 30.19 & 31.78 & 31.39 & 30.71\\
    & Mean &  53.09 & 30.56 & 38.03 & 29.62 & 25.45 & 29.96 & 27.22 & 26.73 & 31.84 & 30.13 & 32.26\\
    \midrule
    \multirow{3}{*}{\makecell{LLaMA-2\\(mntp)}} & EOS & 28.92 & 28.53 & 30.28 & 28.83 & 30.58 & 28.37 & 28.79 & 31.11 & 29.96 & 31.24 & 29.66\\
    & Mean &  30.87 & 29.41 & 31.08 & 28.98 & 27.53 & 29.89 & 28.17 & 26.33 & 29.00 & 28.61 & 28.99\\
    & Weighted Mean & 30.45 & 28.33 & 30.55 & 28.85 & 27.25 & 29.47 & 28.22 & 25.71 & 29.28 & 28.38 & 28.65\\
    \midrule
    \multirow{3}{*}{\makecell{LLaMA-2\\(mntp-unsup-simcse)}} & EOS & 38.17 & 35.56 & 39.46 & 37.97 & 42.25 & 38.12 & 33.04 & 40.98 & 35.59 & 39.13 & 38.03\\
\bottomrule
    \end{tabular}
    }
\end{table}

\begin{table}[h!]
    \centering
    \caption{Average AUPRC(\%) on SST2.}
    \label{tab-AUPR-sst2}
    \resizebox{\textwidth}{!}{
    \begin{tabular}{c|c|c|c|c|c|c|c|c|c|c|c|c}
    \toprule
    \textbf{Embedding} & \textbf{Pooling} & OCSVM & IForest & LOF & PCA & KNN & KDE & ECOD & AE & DSVDD & DPAD & Avg\\
    \midrule
    GloVe & Mean &  48.76 & 47.88 & 60.76 & 47.94 & 51.37 & 48.01 & 47.34 & 51.96 & 57.41 & 56.31 & 51.77\\
    \midrule
    \multirow{2}{*}{BERT} & CLS & 53.33 & 53.00 & 57.23 & 52.81 & 56.73 & 52.85 & 51.49 & 55.72 & 54.52 & 59.24 & 54.69\\
    & Mean & 52.39 & 51.76 & 57.69 & 52.02 & 57.90 & 51.91 & 50.49 & 56.52 & 53.76 & 60.52 & 54.50\\
    \midrule
    \multirow{3}{*}{\makecell{LLaMA-2\\(mntp)}} & EOS &  51.25 & 51.52 & 54.59 & 51.14 & 56.92 & 50.89 & 50.55 & 58.29 & 52.29 & 62.98 & 54.04\\
    & Mean &  50.93 & 50.65 & 55.74 & 50.18 & 55.74 & 50.29 & 49.84 & 57.48 & 52.58 & 61.24 & 53.47\\
    & Weighted Mean &  50.41 & 50.46 & 55.68 & 49.87 & 55.82 & 49.85 & 49.48 & 56.98 & 51.97 & 61.97 & 53.25\\
    \midrule
    \multirow{3}{*}{\makecell{LLaMA-2\\(mntp-unsup-simcse)}} & EOS &  58.39 & 55.96 & 67.03 & 58.43 & 71.81 & 58.41 & 53.16 & 76.37 & 60.25 & 69.90 & 62.97\\
\bottomrule
    \end{tabular}
    }
\end{table}

\begin{table}[h!]
    \centering
    \caption{Average AUPRC(\%) on SMS-SPAM.}
    \label{tab-AUPR-sms_spam}
    \resizebox{\textwidth}{!}{
    \begin{tabular}{c|c|c|c|c|c|c|c|c|c|c|c|c}
    \toprule
    \textbf{Embedding} & \textbf{Pooling} & OCSVM & IForest & LOF & PCA & KNN & KDE & ECOD & AE & DSVDD & DPAD & Avg\\
    \midrule
    GloVe & Mean &  26.52 & 22.75 & 25.21 & 23.05 & 23.86 & 24.57 & 22.44 & 22.00 & 49.75 & 23.83 & 26.40\\
    \midrule
    \multirow{2}{*}{BERT} & CLS &  32.75 & 21.45 & 28.19 & 22.44 & 19.23 & 27.88 & 20.63 & 21.01 & 34.94 & 22.18 & 25.07\\
    & Mean &  28.47 & 25.40 & 30.48 & 26.59 & 21.24 & 26.97 & 20.99 & 23.91 & 74.98 & 33.71 & 31.27\\
    \midrule
    \multirow{3}{*}{\makecell{LLaMA-2\\(mntp)}} & EOS &  56.19 & 48.91 & 41.10 & 50.68 & 22.32 & 49.64 & 30.43 & 28.45 & 25.98 & 35.36 & 38.91\\
    & Mean &  63.63 & 56.71 & 30.73 & 54.21 & 19.45 & 57.92 & 30.47 & 19.16 & 27.26 & 20.75 & 38.03\\
    & Weighted Mean &  71.68 & 65.50 & 43.99 & 64.01 & 19.60 & 68.09 & 33.36 & 20.57 & 24.20 & 22.68 & 43.37\\
    \midrule
    \multirow{3}{*}{\makecell{LLaMA-2\\(mntp-unsup-simcse)}} & EOS &  61.66 & 50.00 & 63.14 & 61.91 & 61.57 & 59.93 & 36.33 & 64.90 & 50.76 & 63.95 & 57.42\\
\bottomrule
    \end{tabular}
    }
\end{table}

\begin{table}[h!]
    \centering
    \caption{Average AUPRC(\%) on Reuters21578.}
    \label{tab-AUPR-reuters}
    \resizebox{\textwidth}{!}{
    \begin{tabular}{c|c|c|c|c|c|c|c|c|c|c|c|c}
    \toprule
    \textbf{Embedding} & \textbf{Pooling} & OCSVM & IForest & LOF & PCA & KNN & KDE & ECOD & AE & DSVDD & DPAD & Avg\\
    \midrule
    GloVe & Mean &  84.40 & 84.32 & 86.11 & 84.53 & 87.87 & 83.97 & 75.58 & 87.37 & 68.23 & 87.91 & 83.03\\
    \midrule
    \multirow{2}{*}{BERT} & CLS &  67.87 & 64.83 & 83.76 & 64.11 & 65.62 & 64.13 & 58.90 & 69.48 & 60.42 & 69.37 & 66.85\\
    & Mean &  82.12 & 82.44 & 92.20 & 81.75 & 87.02 & 79.84 & 71.82 & 87.01 & 82.39 & 90.75 & 83.73\\
    \midrule
    \multirow{3}{*}{\makecell{LLaMA-2\\(mntp)}} & EOS &  76.00 & 76.47 & 69.21 & 77.28 & 73.92 & 76.74 & 73.87 & 75.79 & 76.00 & 77.65 & 75.29\\
    & Mean &  72.41 & 75.13 & 73.37 & 73.73 & 72.38 & 71.49 & 68.02 & 78.75 & 68.66 & 72.81 & 72.68\\
    & Weighted Mean &  72.38 & 75.38 & 72.96 & 73.83 & 72.24 & 71.96 & 68.32 & 73.98 & 69.47 & 73.22 & 72.37\\
    \midrule
    \multirow{3}{*}{\makecell{LLaMA-2\\(mntp-unsup-simcse)}} & EOS &  95.64 & 92.78 & 90.29 & 95.62 & 96.33 & 95.27 & 88.33 & 97.07 & 91.50 & 95.81 & 93.86\\
\bottomrule
    \end{tabular}
    }
\end{table}

\begin{table}[h!]
    \centering
    \caption{Average AUPRC(\%) on IMDB.}
    \label{tab-AUPR-imdb}
    \resizebox{\textwidth}{!}{
    \begin{tabular}{c|c|c|c|c|c|c|c|c|c|c|c|c}
    \toprule
    \textbf{Embedding} & \textbf{Pooling} & OCSVM & IForest & LOF & PCA & KNN & KDE & ECOD & AE & DSVDD & DPAD & Avg\\
    \midrule
    GloVe & Mean &  59.62 & 59.62 & 62.26 & 59.50 & 59.62 & 59.62 & 57.34 & 60.77 & 60.38 & 60.98 & 59.97\\
    \midrule
    \multirow{2}{*}{BERT} & CLS &  59.02 & 59.64 & 59.57 & 58.94 & 60.02 & 59.19 & 56.97 & 59.39 & 62.74 & 60.45 & 59.59\\
    & Mean &  58.52 & 58.70 & 61.10 & 58.24 & 60.29 & 58.07 & 55.39 & 59.87 & 59.81 & 59.88 & 58.99\\
    \midrule
    \multirow{3}{*}{\makecell{LLaMA-2\\(mntp)}} & EOS &  61.30 & 61.22 & 62.38 & 61.30 & 61.69 & 61.36 & 61.09 & 61.60 & 61.64 & 62.11 & 61.57\\
    & Mean &  60.30 & 60.38 & 61.40 & 60.27 & 61.23 & 60.57 & 59.92 & 61.26 & 61.37 & 61.91 & 60.86\\
    & Weighted Mean &  60.53 & 60.52 & 61.84 & 60.47 & 61.33 & 60.78 & 60.14 & 61.32 & 61.36 & 61.96 & 61.02\\
    \midrule
    \multirow{3}{*}{\makecell{LLaMA-2\\(mntp-unsup-simcse)}} & EOS &  59.52 & 59.14 & 65.87 & 58.66 & 64.56 & 58.63 & 54.35 & 65.64 & 61.06 & 62.37 & 60.98\\
\bottomrule
    \end{tabular}
    }
\end{table}

\begin{table}[h!]
    \centering
    \caption{Average AUPRC(\%) on Enron.}
    \label{tab-AUPR-enron}
    \resizebox{\textwidth}{!}{
    \begin{tabular}{c|c|c|c|c|c|c|c|c|c|c|c|c}
    \toprule
    \textbf{Embedding} & \textbf{Pooling} & OCSVM & IForest & LOF & PCA & KNN & KDE & ECOD & AE & DSVDD & DPAD & Avg\\
    \midrule
    GloVe & Mean &  37.93 & 36.93 & 40.82 & 37.03 & 44.10 & 38.10 & 34.42 & 41.41 & 34.71 & 49.17 & 39.46\\
    \midrule
    \multirow{2}{*}{BERT} & CLS &  32.16 & 32.18 & 57.63 & 32.31 & 41.17 & 31.81 & 31.74 & 37.95 & 27.95 & 57.25 & 38.22\\
    & Mean &  31.62 & 31.79 & 57.16 & 31.83 & 35.86 & 31.86 & 30.67 & 34.47 & 34.99 & 52.55 & 37.28\\
    \midrule
    \multirow{3}{*}{\makecell{LLaMA-2\\(mntp)}} & EOS &  34.63 & 34.84 & 44.46 & 34.71 & 36.72 & 35.47 & 33.59 & 39.80 & 32.79 & 65.89 & 39.29\\
    & Mean &  32.36 & 33.56 & 57.62 & 32.55 & 37.46 & 32.05 & 31.93 & 36.79 & 32.99 & 47.38 & 37.47\\
    & Weighted Mean &  32.56 & 33.13 & 56.84 & 32.49 & 35.75 & 32.16 & 31.98 & 36.02 & 33.88 & 47.08 & 37.19\\
    \midrule
    \multirow{3}{*}{\makecell{LLaMA-2\\(mntp-unsup-simcse)}} & EOS &  54.05 & 44.27 & 83.79 & 46.85 & 72.03 & 46.89 & 42.63 & 95.97 & 66.67 & 91.42 & 64.46\\
\bottomrule
    \end{tabular}
    }
\end{table}

\begin{table}[h!]
    \centering
    \caption{Average AUPRC(\%) on 20Newsgroups.}
    \label{tab-AUPR-20newsgroups}
    \resizebox{\textwidth}{!}{
    \begin{tabular}{c|c|c|c|c|c|c|c|c|c|c|c|c}
    \toprule
    \textbf{Embedding} & \textbf{Pooling} & OCSVM & IForest & LOF & PCA & KNN & KDE & ECOD & AE & DSVDD & DPAD & Avg\\
    \midrule
    GloVe & Mean &  13.13 & 13.61 & 14.79 & 13.52 & 15.12 & 13.16 & 13.22 & 13.46 & 12.16 & 14.07 & 13.62\\
    \midrule
    \multirow{2}{*}{BERT} & CLS &  13.26 & 13.82 & 21.37 & 13.90 & 14.39 & 13.44 & 13.63 & 15.17 & 11.60 & 14.50 & 14.51\\
    & Mean &  13.09 & 13.74 & 16.96 & 13.47 & 13.84 & 13.32 & 12.98 & 13.98 & 14.34 & 14.80 & 14.05\\
    \midrule
    \multirow{3}{*}{\makecell{LLaMA-2\\(mntp)}} & EOS &  11.46 & 11.13 & 10.56 & 10.83 & 11.70 & 10.55 & 10.89 & 11.07 & 10.57 & 11.10 & 10.99\\
    & Mean &  13.89 & 12.18 & 12.80 & 12.41 & 11.77 & 12.82 & 12.33 & 11.79 & 11.59 & 11.56 & 12.31\\
    & Weighted Mean &  13.82 & 12.03 & 12.74 & 12.30 & 11.67 & 12.58 & 12.24 & 11.41 & 11.29 & 11.43 & 12.15\\
    \midrule
    \multirow{3}{*}{\makecell{LLaMA-2\\(mntp-unsup-simcse)}} & EOS &  15.61 & 16.01 & 13.98 & 15.43 & 14.46 & 15.19 & 13.84 & 14.16 & 14.10 & 14.11 & 14.69\\
\bottomrule
    \end{tabular}
    }
\end{table}

\end{document}